\documentclass{article}

\usepackage{microtype}
\usepackage{graphicx}
\usepackage{booktabs} % for professional tables
\usepackage{times}
\usepackage{epsfig}
\usepackage{amsmath}
\usepackage{amssymb}
\usepackage{booktabs}
\usepackage{caption}
\usepackage[table,xcdraw,dvipsnames]{xcolor}
\usepackage{multirow}
\usepackage{subcaption}
\usepackage{enumitem}
\usepackage{balance}
\usepackage{overpic}

\makeatletter
\newcommand*\bigcdot{\mathpalette\bigcdot@{.5}}
\newcommand*\bigcdot@[2]{\mathbin{\vcenter{\hbox{\scalebox{#2}{$\m@th#1\bullet$}}}}}
\makeatother

\definecolor{c2}{HTML}{FBD9BD}
\definecolor{c3}{HTML}{fe793d}
\definecolor{c4}{HTML}{eedeb0}
\definecolor{c5}{HTML}{00FFFF}
\definecolor{c6}{HTML}{FF00FF}

\definecolor{rouse}{rgb}{0.981,0.961,0.941}

\usepackage[pagebackref=true,breaklinks=true,letterpaper=true,colorlinks,bookmarks=false]{hyperref}

 \usepackage[accepted]{icml2025}
\usepackage{icml2025}

\usepackage{amsmath}
\usepackage{amssymb}
\usepackage{mathtools}
\usepackage{amsthm}

\usepackage[capitalize]{cleveref}
\crefname{section}{Sec.}{Secs.}
\Crefname{section}{Section}{Sections}
\Crefname{table}{Table}{Tables}
\theoremstyle{plain}

\theoremstyle{definition}

\theoremstyle{remark}

\usepackage[textsize=tiny]{todonotes}

\icmltitlerunning{Submission}

\begin{document}

\twocolumn[
\icmltitle{\textbf{RUN}: Reversible Unfolding Network for Concealed Object Segmentation
% and Beyond
}




\begin{icmlauthorlist}
\icmlauthor{Chunming He}{yyy},
\icmlauthor{Rihan Zhang}{yyy},
\icmlauthor{Fengyang Xiao}{yyy},
\icmlauthor{Chengyu Fang}{comp},
\icmlauthor{Longxiang Tang}{comp},\\
\icmlauthor{Yulun Zhang}{sch},
\icmlauthor{Linghe Kong}{sch},
\icmlauthor{Deng-Ping Fan}{zzz},
\icmlauthor{Kai Li}{eee},
\icmlauthor{Sina Farsiu}{yyy}

\end{icmlauthorlist}

\icmlaffiliation{yyy}{BME, Duke University, Durham, NC, US.}
\icmlaffiliation{comp}{SIGS, Tsinghua University, Shenzhen, Guangdong, China.}
\icmlaffiliation{sch}{MoE Key Lab of Artificial Intelligence, AI Institute, Shanghai Jiao Tong University, Shanghai, China.}
\icmlaffiliation{zzz}{Nankai Institute of Advanced Research (SHENZHEN-FUTIAN), Guangdong, China.}
\icmlaffiliation{eee}{Meta, CA, US.}

%\icmlcorrespondingauthor{Chunming He}{chunming.he@duke.edu}
\icmlcorrespondingauthor{Sina Farsiu}{sina.farsiu@duke.edu}

\icmlkeywords{Machine Learning, ICML}

{\vspace{0.75em}
% \begin{figure*}[h]
	\includegraphics[width=0.99\linewidth]{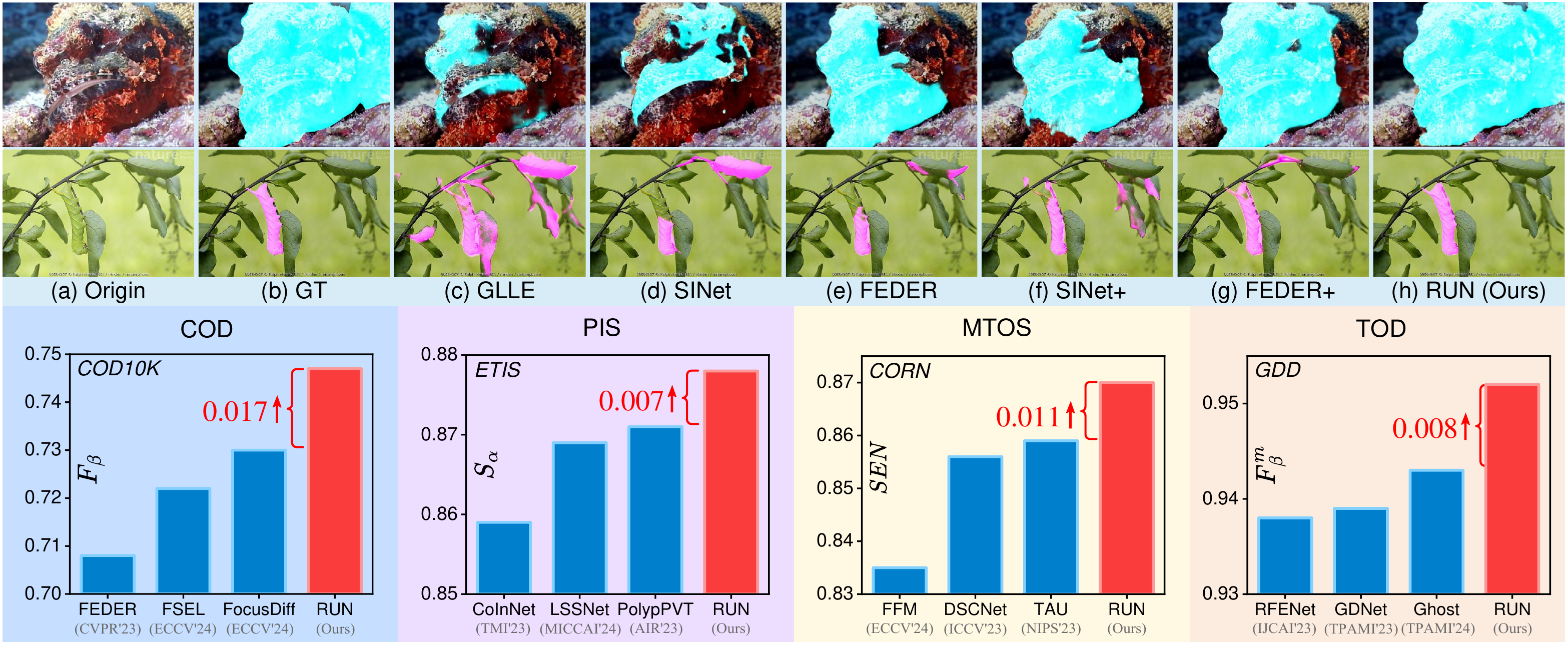}\vspace{-4mm}
    % \put(5,10){\tiny COD}
%     \put(15,75){\textcolor{c5}{\textbf{Blue: Concealed Object Mask}}}
%     \put(15,70){\textcolor{c6}{\textbf{Pink: Concealed Object Mask}}}     
	\captionof{figure}{Results of existing COS methods
%		, including GLLE~\cite{he2019image}, SINet~\cite{fan2020camouflaged}, and FEDER~\cite{He2023Camouflaged}. 
		Our RUN demonstrates superiority in accurately segmenting concealed objects (in the top section) and achieves leading places across multiple COS tasks (in the bottom section): camouflaged object detection (COD), polyp image segmentation (PIS), medical tubular object segmentation (MTOS), and transparent object detection (TOD).
    In the top section, concealed object masks are highlighted in {\textcolor{c5}{blue}} and {\textcolor{c6}{pink}}, overlaid on the original images for visual clarity. FEDER+ and SINet+ indicate integrating FEDER and SINet with our RUN framework. For the bottom section, we employ commonly used datasets, methods, and metrics.
%     relevant to each task.
 }
	\label{fig:Intro}
	\vspace{-5mm}
% \end{figure*}
}

\vskip 0.3in
]



 \printAffiliationsAndNotice{}  % leave blank if no need to mention equal contribution

\begin{abstract}
Existing concealed object segmentation (COS) methods frequently utilize reversible strategies to address uncertain regions. However, these approaches are typically restricted to the mask domain, leaving the potential of the RGB domain underexplored. 
To address this, we propose the Reversible Unfolding Network (RUN), which applies reversible strategies across both mask and RGB domains through a theoretically grounded framework, enabling accurate segmentation.
% we propose a Reversible Unfolding Network (RUN) in this paper to theoretically integrate the information from the mask level and the RGB level, ensuring accurate segmentation. 
RUN first formulates a novel COS model by incorporating an extra residual sparsity constraint to minimize segmentation uncertainties. The iterative optimization steps of the proposed model are then unfolded into a multistage network, with each step corresponding to a stage. Each stage of RUN consists of two reversible modules: the Segmentation-Oriented Foreground Separation (SOFS) module and the Reconstruction-Oriented Background Extraction (ROBE) module. SOFS applies the reversible strategy at the mask level and introduces Reversible State Space to capture non-local information. ROBE extends this to the RGB domain, employing a reconstruction network to address conflicting foreground and background regions identified as distortion-prone areas, which arise from their separate estimation by independent modules. As the stages progress, RUN gradually facilitates reversible modeling of foreground and background in both the mask and RGB domains, directing the network's attention to uncertain regions and mitigating false-positive and false-negative results. Experiments demonstrate our superiority and highlight the potential of unfolding-based frameworks for COS and other high-level vision tasks. Code will be released at https://github.com/ChunmingHe/RUN.
\end{abstract}
\setlength{\abovedisplayskip}{2pt}
\setlength{\belowdisplayskip}{2pt}
\begin{figure*}[h]
\setlength{\abovecaptionskip}{0cm}
	\centering
	\includegraphics[width=\linewidth]{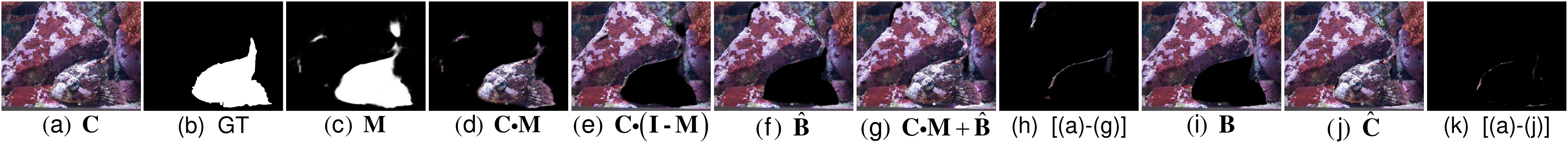}\vspace{-4mm}
	\caption{Correspondence between uncertainties in the mask domain and distortions in the RGB domain.
    % Results from the $2^{nd}$ stage are presented as an example. 
    ${\mathbf{C}}$ is the concealed image and $\hat{\mathbf{B}}$ is the estimated background, which has conflicting judgments of concealed regions with the mask $\mathbf{M}$. This conflict leads to distortion-prone areas in their direct combination (g). Panel (h) illustrates the difference between (g) and the original image (a). However, after refinement through the network $\mathcal{B}(\bigcdot)$, the reconstructed image $\hat{\mathbf{C}}$ becomes much closer to the original image, accompanied by a refined background $\mathbf{B}$ with improved accuracy. This refined background is passed to the next stage to further facilitate segmentation.
    }
	\label{fig:Recon}
	\vspace{-3.5mm}
\end{figure*}
\section{Introduction}

Concealed object segmentation (COS) aims to segment objects that are visually blended with their surroundings. It serves as an umbrella term with various applications, including camouflaged object detection~\cite{he2023strategic}, polyp image segmentation~\cite{he2023weaklysupervised}, and transparent object detection~\cite{xiao2023concealed}, among others.

COS is a challenging problem due to the intrinsic similarity between the object and its background. Traditional methods address this challenge by relying on manually designed models with hand-crafted feature extractors tailored to subtle differences in textures, intensities, and colors~\cite{he2019image}. While offering clear interpretability, they often structure in complex scenarios. Deep learning advances COS by leveraging its strong generalization capabilities, driven by powerful feature extraction mechanisms. Early learning-based approaches, such as SINet~\cite{fan2020camouflaged}, primarily focus on foreground regions for segmentation, often overlooking discriminative cues in the background, leading to suboptimal performance (see~\cref{fig:Intro}). Recent algorithms, such as FEDER~\cite{He2023Camouflaged}, have sought to refine segmentation masks by reversibly modeling both foreground and background regions at the mask-level.

Reversible modeling enhances the network’s capacity to extract subtle discriminative cues by directing attention to uncertain regions—pixels with values that are neither 1 nor 0—thus improving segmentation results. However, current methods restrict the application of reversible strategies to the mask level, leaving the potential of the RGB domain underexplored. Such information can assist in identifying discriminative cues and enhancing segmentation quality. As shown in~\cref{fig:Recon} (d) and (e), when reversibly separating the image into foreground and background regions based on the mask, the uncertainty regions in the mask tend to manifest as color distortion in the RGB space. Addressing these translates to a more precise separation of the foreground and background.
% Removing those distortions also means more accurately separate foreground and background. 
In this case, two seemingly independent tasks—object segmentation and distortion restoration—share the same optimization goal. Existing research has shown that jointly optimizing such tasks helps guide the network toward an optimal solution~\cite{he2023HQG}.

To achieve this, we first introduce a deep unfolding network termed the Reversible Unfolding Network (RUN) for COS. RUN established a theoretical foundation to reversibly integrate the two aforementioned tasks, rather than directly combining them, to achieve more accurate segmentation.
The COS task is formulated as a foreground-background separation process, and a new segmentation model is developed by incorporating a residual sparsity constraint to reduce segmentation uncertainties. The iterative optimization steps of the model-based solution are then unfolded into a multi-stage network, with each step corresponding to a stage. 
Each stage comprises two reversible modules: the Segmentation-Oriented Foreground Separation (SOFS) module and the Reconstruction-Oriented Background Extraction (ROBE) module. By integrating optimization solutions with deep networks, our RUN framework achieves an effective balance between interpretability and generalizability.

We implement the reversible strategy within the mask domain in SOFS and within the RGB domain in ROBE. 
In SOFS, the mask is initially updated strictly according to the optimization solution. Subsequently, the Reversible State Space (RSS) module, recognized for its strong capacity to extract non-local information, is employed to refine the segmentation mask using the previously estimated mask and background. 
In ROBE, the process begins with a mathematical update of the background. A lightweight network is then used to reconstruct the entire image, while simultaneously refining the background, based on the estimated foreground and background. 
Since the estimation of foreground and background regions is performed by distinct modules, their assessments of concealed content can differ (see~\cref{fig:Recon} (d) and (f)). Regions of conflicting judgments are identified as distortion-prone areas during the reconstruction process (see~\cref{fig:Recon} (g) and (h)). This auxiliary reconstruction task, which targets the resolution of such distortions, effectively directs the network’s attention to challenging regions where distinguishing between foreground and background is particularly difficult, improving segmentation performance.

% Given that separate modules are used to estimate foreground and background regions, their judgments regarding concealed content can be distinct. 
% Regions of conflicting judgments are identified as distortion-prone regions during image reconstruction. 
% Hence, this auxiliary task, aiming to remove those distortions, directs the network’s attention to challenging regions where distinguishing foreground from background is particularly difficult, thereby facilitating the segmentation task.

% RUN enables reversible modeling of foreground and background in both the mask and RGB domains and thus effectively reduces false-positive and false-negative regions as the stages progress. 

As the stages progress, RUN incrementally facilitates reversible modeling of foreground and background in both the mask and RGB domains. This approach effectively focuses the network on uncertain regions, reducing false-positive and false-negative outcomes. Notably, RUN exhibits high flexibility, allowing seamless integration with existing methods to achieve further performance enhancements.

Our contributions are summarized as follows:

(1) We propose RUN for the COS task. To the best of our knowledge, this represents the first application of a deep unfolding network to address the COS problem, thereby balancing interpretability and generalizability.

(2) RUN proposes a novel COS model designed to reduce segmentation uncertainties and introduces SOFS and ROBE modules to integrate model-based optimization solutions with deep networks. By enabling reversible modeling of foreground and background across both the mask and RGB domains, RUN directs the network's focus to uncertain regions, reducing false-positive and false-negative outcomes.

% RUN enables reversible modeling of foreground and background across both mask and RGB domains, directing the network's attention to uncertain regions and mitigating false-positive and false-negative outcomes.

(3) Experiments on five COS tasks, as well as salient object detection, validate the superiority of our RUN method. Besides, its plug-and-play structure underscores the effectiveness and adaptability of unfolding-based frameworks for the COS task and other high-level vision tasks.

% (3) Experiments across five COS tasks, as well as salient object detection, demonstrate the superiority of our RUN method. Besides, the framework exhibits strong potential as a plug-and-play module for enhancing the performance of existing methods. This underscores the effectiveness and versatility of unfolding-based frameworks for COS and other high-level vision tasks.

\section{Related Works}
\noindent \textbf{Concealed object segmentation}. Deep learning methods have advanced COS~\cite{xiao2024survey}. 
Among them, those using reversible techniques to segment from foreground and background aspects are gaining attention.
PraNet~\cite{fan2020pranet} introduced a parallel structure with reversible attention to enhance segmentation. FEDER~\cite{He2023Camouflaged}
% added an edge detection task and 
used foreground and background masks to identify concealed objects with edge assistance. 
BiRefNet~\cite{zheng2024bilateral} proposed a reconstruction module to refine the mask with gradient information.
However, they only focus on the mask level, leaving the RGB domain underexplored. Hence, we propose the first deep unfolding network, RUN, for COS. RUN proposes a novel COS model and introduces SOFS and ROBE. By integrating optimization solutions with deep networks, RUN enables reversible modeling across mask and RGB domains, improving segmentation accuracy.

\noindent \textbf{Deep unfolding network}. The deep unfolding network, a well-established technique in low-level vision, integrates model-based and learning-based approaches~\cite{he2023degradation,fang2024real}, offering enhanced interpretability compared to pure learning-based methods. However, its application in high-level vision remains underexplored, primarily due to the lack of explicit intrinsic models for high-level vision tasks.
In this paper, we introduce a deep unfolding network, RUN, in COS and formulate a novel COS model. RUN achieves more accurate segmentation results by integrating optimization-based solutions with deep networks, verifying its potential for advancing COS.

% \begin{algorithm}[tb]
%    \caption{Bubble Sort}
%    \label{alg:example}
% \begin{algorithmic}
%    \STATE {\bfseries Input:} data $x_i$, size $m$
%    \REPEAT
%    \STATE Initialize $noChange = true$.
%    \FOR{$i=1$ {\bfseries to} $m-1$}
%    \IF{$x_i > x_{i+1}$}
%    \STATE Swap $x_i$ and $x_{i+1}$
%    \STATE $noChange = false$
%    \ENDIF
%    \ENDFOR
%    \UNTIL{$noChange$ is $true$}
% \end{algorithmic}
% \end{algorithm}

\section{Methodology}
% In this section, we introduce the proposed novel COS model in~\cref{Sec:COSM}, optimize it using the proximal gradient algorithm in~\cref{Sec:Optimization}, and unfold the iterative steps of the traditional method into a multi-stage network, RUN, in~\cref{Sec:DUN}.
\vspace{-1mm}
\subsection{COS Model}~\label{Sec:COSM}\vspace{-1mm}
A concealed image $\mathbf{C}$ can be decomposed into its foreground region $\mathbf{F}$ and background region $\mathbf{B}$, expressed as
\begin{equation}\label{Eq:Separation}
\mathbf{C}=\mathbf{F}+\mathbf{B}.
\end{equation}
Based on~\cref{Eq:Separation}, the foreground and background regions can be obtained by optimizing the objective function:
\begin{equation}\label{Eq:BasicModel}
   L\left(\mathbf{F}, \mathbf{B}\right)=\frac{1}{2}\|\mathbf{C}-\mathbf{F}-\mathbf{B}\|_2^2 + \beta\varphi(\mathbf{F}) + \lambda \phi(\mathbf{B}),
\end{equation}
where $\|\bigcdot\|_2$ is a $\ell_2$-norm for smooth. $\varphi(\mathbf{F})$ and $\phi(\mathbf{B})$ are regularization terms for $\mathbf{F}$ and $\mathbf{B}$ with two trade-off parameters $\beta$ and $\lambda$. 
To suit the segmentation task, we directly focus on the mask $\mathbf{M}$, where $\mathbf{F}= \mathbf{C}\bigcdot\mathbf{M}$, and $\bigcdot$ is dot product. Substituting into \cref{Eq:BasicModel}, the objective function becomes
\begin{equation}\label{Eq:BasicModel1}
   L\left(\mathbf{M}, \mathbf{B}\right)=\frac{1}{2}\|\mathbf{C}- \mathbf{C}\bigcdot\mathbf{M} -\mathbf{B}\|_2^2 + \mu\psi(\mathbf{M}) + \lambda \phi (\mathbf{B}),
\end{equation}
where $\psi(\mathbf{M})$ and $\mu$ are the regularization term and trade-off parameter for $\mathbf{M}$. Due to the intrinsic ambiguity of foreground objects in concealed images and the diverse nature of their backgrounds, manually defining regularization terms for $\mathbf{M}$ and $\mathbf{B}$ can be challenging. To address this, we utilize deep neural networks to implicitly learn these constraints in a data-driven manner.
% Due to the nature of concealed images, their foreground objects often exhibit ambiguous structures, while their background can vary significantly. These challenges make it difficult to enforce manually designed regularizations on $\mathbf{M}$ and $\mathbf{B}$. To address this, we employ deep networks to implicitly learn these constraints in a data-driven manner, rather than relying on explicitly designed regularization terms. 
Beyond the two intrinsic regularization terms above, we introduce an extra residual sparsity constraint $\mathcal{S}(\bigcdot)$ to further refine segmentation and minimize uncertainties. This leads to the final objective function:
\begin{equation}\label{Eq:FinalModel}
\begin{aligned}
       L\left(\mathbf{M}, \mathbf{B}\right)&=\frac{1}{2}\|\mathbf{C}- \mathbf{C}\bigcdot\mathbf{M} -\mathbf{B}\|_2^2 + \mu\psi(\mathbf{M})\\
       & + \lambda \phi (\mathbf{B})
       +\alpha\mathcal{S}\left(\mathbf{w}\bigcdot\left(\mathbf{M}-\widetilde{\mathbf{M}}\right)\right),
\end{aligned}
\end{equation}
where $\alpha$ controls the weight of the sparsity constraint, $\mathcal{S}(\bigcdot)$ represents an $\ell_1$-norm, $\widetilde{\mathbf{M}}$ is the refined mask after uncertainty-removal mapping, and $\mathbf{w}$ is the attention map. For a pixel located at $(i,j)$, $\widetilde{\mathbf{M}}$ and $\mathbf{w}$ can be defined as 
\begin{equation}
\begin{aligned}
    \widetilde{\mathbf{M}}_{(i,j)}=&
\left\{
\begin{array}{ll}
    0.1  &\mathbf{M}_i \in [0.1,0.4),\\
    0.9  &\mathbf{M}_i \in (0.6,0.9],\\
    \mathbf{M}_i & \text{Otherwise}.
\end{array}\right. \\
\mathbf{w}_{(i,j)}=&
\left\{
\begin{array}{ll}
    0  &\mathbf{M}_i \in [0.4,0.6],\\
    1  & \text{Otherwise}.
\end{array}\right.
\end{aligned}
\end{equation}
This design encourages the generation of segmentation masks with high certainty. Following the practice of~\cite{he2023weaklysupervised}, pixels with values in the ambiguous range $[0.4, 0.6]$ are excluded from further consideration, while extreme values for $\widetilde{\mathbf{M}}$ are set to 0.1 and 0.9 instead of 0 and 1 to allow greater flexibility for optimization.

\begin{figure*}[h]
\setlength{\abovecaptionskip}{0cm}
	\centering
	\includegraphics[width=\linewidth]{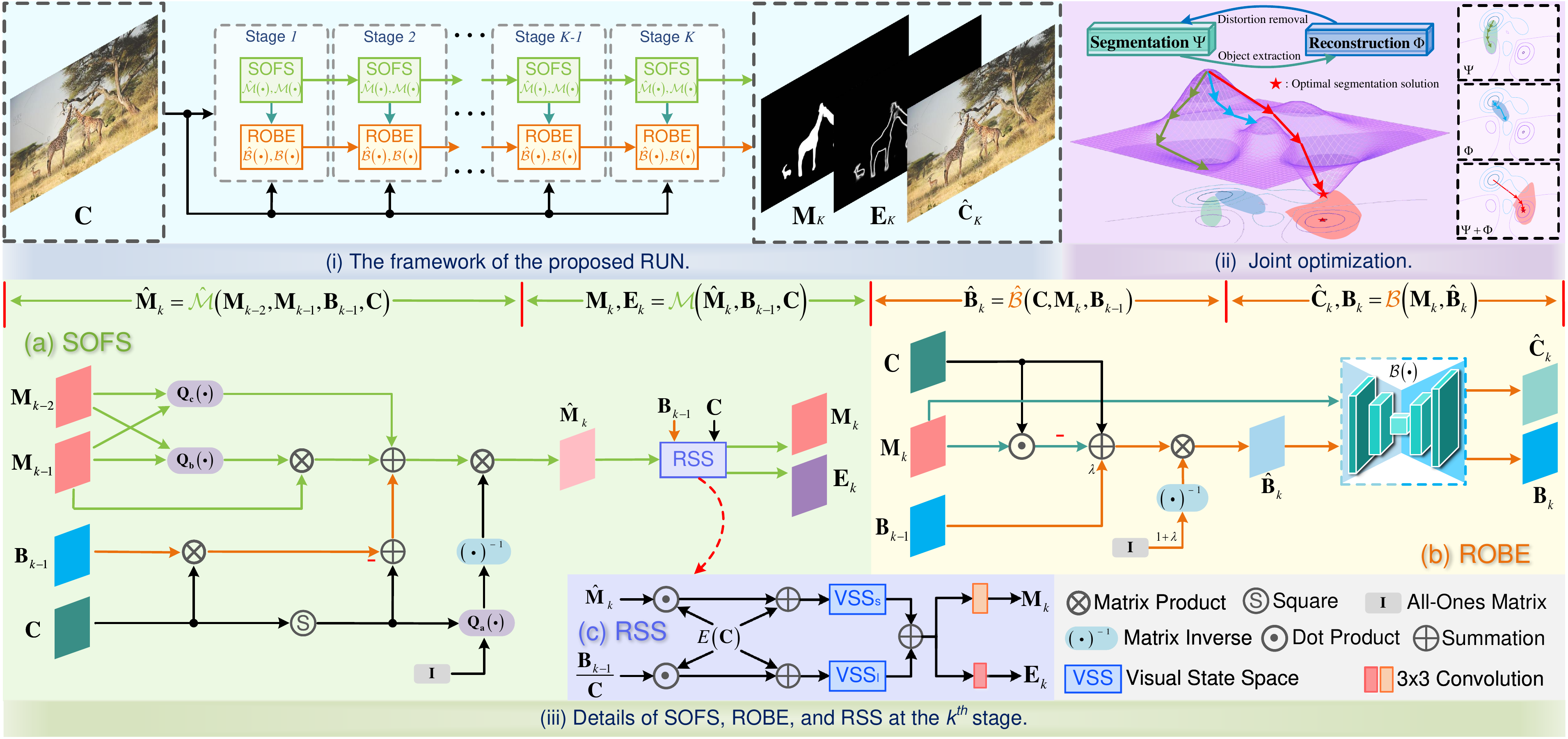}\vspace{-4mm}
	\caption{Framework of our RUN. The network connections in $\hat{\mathcal{M}}(\cdot)$ and $\hat{\mathcal{B}}(\cdot)$ are derived strictly based on mathematical principles, thus enhancing interpretability. For clarity, we replace certain redundant details with $\mathbf{Q}_\mathbf{a}$, $\mathbf{Q}_\mathbf{b}$, and $\mathbf{Q}_\mathbf{c}$ and present $\hat{\mathcal{M}}(\cdot)$ according to~\cref{Eq:SOFSMhat}. The detailed connection can be seen in~\cref{fig:SOFS_Details} in the Appendix. Panel (ii) illustrates that the joint optimization of image segmentation and reconstruction tasks facilitates the network's progression toward an optimal solution.
    % Framework of our RUN. The network connections in $\hat{\mathcal{M}}(\bigcdot)$ and $\hat{\mathcal{B}}(\bigcdot)$ strictly follow mathematical derivation, improving interpretability. We omit some redundant details for clarity. The content in (ii) indicates that jointly optimizing image segmentation and reconstruction tasks better promotes the network toward an optimal solution.
    }
	\label{fig:Framework}
	\vspace{-5mm}
\end{figure*}

\subsection{RUN}\vspace{-1mm}
\subsubsection{Model Optimization}~\label{Sec:Optimization}\vspace{-1mm}
We utilize the proximal gradient algorithm~\cite{fang2024real} to optimize~\cref{Eq:FinalModel}, ultimately deriving the optimal mask $\mathbf{M}^*$ and background $\mathbf{B}^*$:
\begin{equation}
   \left\{\mathbf{M}^*,\mathbf{B}^*\right\} = \arg \underset{\mathbf{M}, \mathbf{B}}{\min } \ 
   L\left(\mathbf{M}, \mathbf{B}\right).
\end{equation}
% The optimization process alternates between updating the two variables iteratively. Below, we describe the procedure for the $k^{th}$ iteration ($1 \leq k \leq K$).
The optimization process involves alternating updates of the two variables over iterations. Here we take the $k^{th}$ stage ($1\leq k\leq K$) to present the alternative solution process.

\noindent\textbf{Optimizing $\mathbf{M}_k$}. First, the optimization function is partitioned to update the foreground mask $\mathbf{M}_k$:
\begin{equation}\label{Eq:SolutionM1}
\begin{aligned}
       \mathbf{M}_k = \arg \underset{\mathbf{M}}{\min } \ &\frac{1}{2}\|\mathbf{C}- \mathbf{C}\bigcdot\mathbf{M} -\mathbf{B}_{k-1}\|_2^2 + \mu\psi(\mathbf{M})\\
       & +\alpha\mathcal{S}\left(\mathbf{w}_k\bigcdot\left(\mathbf{M}-\widetilde{\mathbf{M}}_k\right)\right).
\end{aligned}
\end{equation} 
The solution comprises two terms: the gradient descent term and the proximal term. To address the proximal term, we introduce an auxiliary variable $\hat{\mathbf{M}}_k$, resulting in:
% The proximal term is formulated by introducing an auxiliary variable $\hat{\mathbf{M}}_k$:
\begin{equation}\label{Eq:SolutionM2}
       \mathbf{M}_k = \text{prox}_\psi\left(\mathbf{B}_{k-1},\hat{\mathbf{M}}_k\right),
\end{equation} 
where $\mathbf{B}_0$ is initialized as zero. Having gotten~\cref{Eq:SolutionM1,Eq:SolutionM2}, $\hat{\mathbf{M}}_k$ can be solved by optimizing:
\begin{equation}\hspace{-2mm}\label{Eq:SolutionM3}
\begin{aligned}
       \hat{\mathbf{M}}_k = &\frac{1}{2}\|\mathbf{C}- \mathbf{C}\bigcdot \hat{\mathbf{M}} -\mathbf{B}_{k-1}\|_2^2 + \frac{\mu}{2}\|\hat{\mathbf{M}}-\mathbf{M}_{k-1}\|_2^2\\
       & +\alpha\mathcal{S}\left(\mathbf{w}_k\bigcdot\left(\hat{\mathbf{M}}-\widetilde{\mathbf{M}}_k\right)\right),
\end{aligned}
\end{equation} 
where $\mathbf{M}_0$ is also initialized as zero. Note that $\mathbf{w}_k$ and $\widetilde{\mathbf{M}}_k$ are constructed based on ${\mathbf{M}_{k-1}}$.
For the term $\mathcal{S}(\mathbf{w}_k\bigcdot(\hat{\mathbf{M}}-\widetilde{\mathbf{M}}_k))$, we employ a Taylor expansion rather than soft thresholding
% the $\ell_1$-norm 
for flexibility in problem-solving. Following the practice of~\cite{goldstein1977optimization}, we approximate $\mathcal{S}(\mathbf{w}_k\bigcdot(\hat{\mathbf{M}}-\widetilde{\mathbf{M}}_k))$ at the $k-1^{th}$ iteration (for simplicity, we let $\mathbf{R}=\mathbf{w}_k\bigcdot(\hat{\mathbf{M}}-\widetilde{\mathbf{M}}_k)$), expressed as follows:
\begin{equation}\hspace{-2mm}\label{Eq:SolutionM4}
      \mathcal{S}\left(\mathbf{R}\right) \approx \dot{\mathcal{S}}\left(\mathbf{R,R_{k-1}}\right), 
\end{equation} 
where 
\begin{equation}\hspace{-2mm}\label{Eq:SolutionM5}
 \dot{\mathcal{S}}(\mathbf{R,R_{k-1}}\!) \!\leftarrow\! \frac{L_{\mathcal{S}}}{2}\|\mathbf{R}\!-\!\mathbf{R}_{k-1} \!+\! \frac{1}{L_{\mathcal{S}}}\nabla\mathcal{S}\!\left(\mathbf{R}_{k-1}\!\right)\!\|_2^2\!+\!C_\mathcal{S},
 % \vspace{-2mm}
\end{equation}
where $L_{\mathcal{S}}$ is the Lipschitz constant. $\nabla\mathcal{S}(\mathbf{R}_{k-1})$ is the Lipschitz continuous gradient function of $\mathcal{S}(\mathbf{R}_{k-1})$ with $C_\mathcal{S}$, a positive constant that can be omitted in optimization. 
Substituting into~\cref{Eq:SolutionM3}, we obtain the following equations:
\begin{equation}\hspace{-3mm}\label{Eq:SolutionM6}
\begin{aligned}
       \hat{\mathbf{M}}_k = &\frac{1}{2}\|\mathbf{C}- \mathbf{C}\bigcdot \hat{\mathbf{M}} -\mathbf{B}_{k-1}\|_2^2 + \frac{\mu}{2}\|\hat{\mathbf{M}}-\mathbf{M}_{k-1}\|_2^2\\
       & +\frac{\alpha L_{\mathcal{S}}}{2}\|\mathbf{R}\!-\!\mathbf{R}_{k-1} \!+\! \frac{1}{L_{\mathcal{S}}}\nabla\mathcal{S}\left(\mathbf{R}_{k-1}\right)\|_2^2.
\end{aligned}
\end{equation} 
Unlike $\ell_1$-norm methods,~\cref{Eq:SolutionM6} can be solved directly by equating its derivative to zero. The closed-form solution is 
% Without requiring the simulation of soft thresholding like $\ell_1$-norm~\cite{wu2024rpcanet}, \cref{Eq:SolutionM6} can be easily solved by equating the derivative to zero.  The closed-form solution is 
\begin{equation}\hspace{-3mm}\label{Eq:SolutionMFinal}
% \begin{aligned}
       \hat{\mathbf{M}}_k \!= \!\left(\mathbf{Q_a}\right)^{-1} \!\left(\mathbf{Q_b}{\mathbf{M}}_{k-1} \!+\! \mathbf{C}^2 \!-\!\mathbf{C B_{k-1}} \!+\! \mathbf{Q_c}\right)\!,
% \end{aligned}
\end{equation} 
where $\mathbf{Q_a}=\mathbf{C}^2+L_{\mathcal{S}}\mathbf{w}_k^2 +\mu \mathbf{I}$, $\mathbf{I}$ is an all-ones matrix, $\mathbf{Q_b}=\alpha L_{\mathcal{S}}\mathbf{w}_k \mathbf{w}_{k-1}+\mu \mathbf{I}$, 
% and $\mathbf{Q_c}=\alpha L_{\mathcal{S}}\mathbf{w}_k (\mathbf{w}_k \widetilde{\mathbf{M}}_k - \mathbf{w}_{k-1} \widetilde{\mathbf{M}}_{k-1}) - \alpha \mathbf{w}_k \nabla \mathcal{S}(\mathbf{w}_{k-1} {\mathbf{M}}_{k-1}-\mathbf{w}_{k-1} \widetilde{\mathbf{M}}_{k-1})$. 
$\mathbf{Q_c}=\alpha L_{\mathcal{S}}\mathbf{w}_k (\mathbf{w}_k\bigcdot \widetilde{\mathbf{M}}_k - \mathbf{Q_d}) - \alpha \mathbf{w}_k \nabla \mathcal{S}(\mathbf{w}_{k-1}\bigcdot {\mathbf{M}}_{k-1}-\mathbf{Q_d})$, and $\mathbf{Q_d}=\mathbf{w}_{k-1}\bigcdot \widetilde{\mathbf{M}}_{k-1}$. 
% $\mathbf{I}$ is an all-ones metrics. 
% Notably, $\mathbf{w}_k$ and $\widetilde{\mathbf{M}}_k$ are derived from ${\mathbf{M}}_{k-1}$, while $\mathbf{w}_{k-1}$ and $\widetilde{\mathbf{M}}_{k-1}$ are based on ${\mathbf{M}}_{k-2}$. 
% For simplicity, no additional notation is introduced to further elaborate.

\noindent\textbf{Optimizing $\mathbf{B}_k$}. The optimization function of $\mathbf{B}_k$ is
\begin{equation}\label{Eq:SolutionB1}
       \mathbf{B}_k=\arg \underset{\mathbf{B}}{\min } \frac{1}{2}\|\mathbf{C}- \mathbf{C}\bigcdot\mathbf{M}_k -\mathbf{B}\|_2^2 + \lambda \phi (\mathbf{B}).
\end{equation}
Same as the optimization rule for $\mathbf{M}_k$, the gradient descent term and the proximal term are correspondingly defined as:
\begin{equation}\hspace{-2mm}\label{Eq:SolutionB2}
\setlength{\abovedisplayskip}{0pt}
\setlength{\belowdisplayskip}{0pt}
       \hat{\mathbf{B}}_k= \frac{1}{2}\|\mathbf{C}- \mathbf{C}\bigcdot\mathbf{M}_k - \hat{\mathbf{B}}\|_2^2 + \frac{\lambda}{2} \|\hat{\mathbf{B}} - \mathbf{B}_{k-1}\|^2_2,
\end{equation} 
\begin{equation}\label{Eq:SolutionB3}
\setlength{\abovedisplayskip}{0pt}
\setlength{\belowdisplayskip}{-5pt}
       \mathbf{B}_k = \text{prox}_\phi (\hat{\mathbf{B}}_{k},\mathbf{M}_k ).
\end{equation} 
The closed-form solution of $\hat{\mathbf{B}}_k$ can be acquired similarly:
\begin{equation}\hspace{-2mm}\label{Eq:SolutionBFinal}
       \hat{\mathbf{B}}_k= \left(\left(1+\lambda \right)\mathbf{I}\right)^{-1}\left(\lambda\mathbf{B}_{k-1}+\mathbf{C}- \mathbf{C}\bigcdot\mathbf{M}_k \right).
\end{equation} 

\subsubsection{Deep Unfolding Mechanism}~\label{Sec:DUN}
We unfold the iterative optimization steps of the model-based solution into a multi-stage network, termed Reversible Unfolding Network (RUN), with each step corresponding to a stage. As shown in~\cref{fig:Framework}, each stage has two reversible modules: the Segmentation-Oriented Foreground Separation (SOFS) and Reconstruction-Oriented Background Extraction (ROBE) modules. 

\noindent\textbf{SOFS}. SOFS, derived from~\cref{Eq:SolutionM2,Eq:SolutionMFinal}, utilizes $\hat{\mathcal{M}}(\bigcdot)$ and $\mathcal{M}(\bigcdot)$ to compute the optimization solution $\hat{\mathbf{M}}$ and the refined mask $\mathbf{M}$ at each stage, respectively. Given $\mathbf{B}_{k-1}$, $\mathbf{M}_{k-1}$, and $\mathbf{M}_{k-2}$, we define $\hat{\mathbf{M}}_k$ as follows:
\begin{equation}\hspace{-3mm}\label{Eq:SOFSMhat}
\begin{aligned}
       \hat{\mathbf{M}}_k &=\! \hat{\mathcal{M}}\left(\mathbf{B}_{k-1}, \mathbf{M}_{k-1}, \mathbf{M}_{k-2}, \mathbf{C}\right)\!, \\ &=\!\left(\mathbf{Q_a}\right)^{-1} \!\left(\mathbf{Q_b}{\mathbf{M}}_{k-1} \!+\! \mathbf{C}^2 \!-\!\mathbf{CB_{k-1}} \!+\! \mathbf{Q_c}\right)\!.
\end{aligned}
\end{equation} 
\cref{Eq:SOFSMhat} retains the same formulation as~\cref{Eq:SolutionMFinal}, but all originally fixed parameters, including $\nabla \mathcal{S}(\bigcdot)$, are relaxed to be learnable, improving the model's generalizability.

To refine the initial mask $\hat{\mathbf{M}}_k$, we introduce the Reversible State Space (RSS) module $RSS(\bigcdot)$, which has a robust capacity for non-local information extraction. 
The RSS module incorporates two Visual State Space (VSS) $VSS(\bigcdot)$ modules~\cite{Liu2024vmamba} with distinct perception fields. The VSS with a small perception field locally refines uncertain regions along the edges from the foreground perspective, while the VSS with a large perception field globally identifies missed segmented regions from the background perspective. This dual-perception mechanism ensures both accurate and comprehensive segmentation results. Following~\cite{He2023Camouflaged}, we also integrate an auxiliary edge output $\mathbf{E}_k$ to further enhance segmentation performance. Consequently, the computation of $\mathbf{M}_k$ and $\mathbf{E}_k$ is defined as:
\begin{equation}\hspace{-3mm}
\begin{aligned}
    \mathbf{M}_k,\mathbf{E}_k &\!=\!  \mathcal{M}( \mathbf{B}_{k-1}\!, \hat{\mathbf{M}}_k,\mathbf{C}) \!= \! RSS ( \mathbf{B}_{k-1},\hat{\mathbf{M}}_k,\mathbf{C}), \\
    &\!=\! conv3(VSS_s(E(\mathbf{C})\bigcdot\hat{\mathbf{M}}_k + E(\mathbf{C})) \\
    &\!+\! VSS_l (E(\mathbf{C})\bigcdot(\mathbf{B}_{k-1}/\mathbf{C}) + E(\mathbf{C}) )),
\end{aligned}
\end{equation}
% \begin{equation}\hspace{-3mm}
% \begin{aligned}
%     RSS ( \mathbf{B}_{k-1},\hat{\mathbf{M}}_k,\mathbf{C}) &=  conv3(VSS_s(E(\mathbf{C})\bigcdot\hat{\mathbf{M}}_k + E(\mathbf{C}))  \\
%      &+ VSS_l (E(\mathbf{C})\bigcdot\mathbf{B}_{k-1} + E(\mathbf{C}) )),
% \end{aligned}
% \end{equation}
where $conv3$ is $3\times3$ convolution. $VSS_s(\bigcdot)$ and $VSS_l(\bigcdot)$ have
% are the VSS modules~\cite{Liu2024vmamba} with 
small and large perception fields, incorporating convolutions with varying kernel sizes. For brevity, we omit the detailed description of VSS.
Unlike low-level vision tasks, segmentation tasks, particularly inherently complex COS, strongly depend on semantic information. It is challenging to extract this fully using a shallow network. To address this, we adopt the common practice of leveraging deep features $E(\mathbf{C})$, extracted from an encoder (default: ResNet50~\cite{he2016deep}). Rather than directly processing the concealed image, this approach enables the extraction of subtle discriminative features, achieving accurate segmentation.

\noindent\textbf{ROBE}. In ROBE, the calculation of $\hat{\mathbf{B}}_k$ relies on $\hat{\mathcal{B}}(\bigcdot)$, similar to~\cref{Eq:SolutionB2} but with the fixed parameters made learnable: 
\begin{equation}\hspace{-2mm}\label{Eq:SOFSBhat}
\begin{aligned}
    \hat{\mathbf{B}}_k&=\hat{\mathcal{B}}\left(\mathbf{B}_{k-1}, \mathbf{M}_k, \mathbf{C}\right), \\
    &= \left(\left(1+\lambda \right)\mathbf{I}\right)^{-1}\left(\lambda\mathbf{B}_{k-1}+\mathbf{C}- \mathbf{C}\bigcdot\mathbf{M}_k \right).
\end{aligned}
\end{equation} 
This is essentially a dynamic fusion of the previously estimated background and the reversed foreground derived in the current stage.
To refine $\hat{\mathbf{B}}_k$, we propose a simple U-shaped network~\cite{he2023HQG} with three layers, denoted as $\mathcal{B}(\bigcdot)$. However, as shown in~\cref{fig:Recon}, since separate modules estimate the foreground and background, their interpretations of the concealed content may differ. Hence, regions with conflicting interpretations are identified as distortion-prone areas in reconstruction. To address this, the network also generates a reconstructed result $\hat{\mathbf{C}}_k$, formulated as:
\begin{equation}\hspace{-2mm}
\begin{aligned}
    {\mathbf{B}}_k, \hat{\mathbf{C}}_k = {\mathcal{B}}\left(\hat{\mathbf{B}}_{k}, \mathbf{M}_k\right).
\end{aligned}
\end{equation} 
$\hat{\mathbf{C}}_k$ is designed to be consistent with the concealed image, thereby mitigating distortions. This alignment fosters consistent judgments between SOFS and ROBE for foreground-background separation, improving segmentation accuracy.
As the stages progress, RUN incrementally facilitates reversible modeling of the foreground and background in both the mask and RGB domains. This iterative process directs the network's attention to regions of uncertainty, reducing false-positive and false-negative outcomes. Hence, RUN ensures robust and accurate segmentation performance.

\noindent\textbf{Loss function}. The loss function comprises a segmentation term and a reconstruction term. We adopt the training strategy from FEDER~\cite{He2023Camouflaged} for the segmentation part. A mean square error loss governs the reconstruction component. The overall loss function is defined as 
\begin{equation}\hspace{-3mm}
\begin{aligned}
		L_t\!&=\!\sum_{k=1}^{K}\frac{1}{2^{K-k}}\!\left[L^w_{BCE}\!\left(\mathbf{M}_k,GT_s\right)\!+\!L^w_{IoU}\!\left(\mathbf{M}_k,GT_s\right)\right.\\
		& +\! L_{dice}\left(\mathbf{E}_k,GT_e\right)+ \|\hat{\mathbf{C}}_k-\mathbf{C} \|^2_2 \ ],
	\end{aligned}
\end{equation}
where $K$ is the number of stages. $L_{BCE}^w$ is the weighted binary cross-entropy loss, $L_{IoU}^w$ is the weighted intersection-over-union loss, and $L_{dice}$ is the dice loss. $GT_s$ and $GT_e$ are the ground truth of the segmentation mask and edge.

\begin{table*}[tbp!]
		\setlength{\abovecaptionskip}{0cm} 
		\setlength{\belowcaptionskip}{-0.2cm}
		\centering
            \caption{Results on camouflaged object detection. SegMaR-1/-4 are SegMaR with one or four stages. 
            % Swin denotes Swin Transformer.
            % ~\cite{liu2021swin}. 
            The best results are marked in \textbf{bold}. For the ResNet50 backbone in the common setting, the best two results are in {\color[HTML]{FF0000} \textbf{red}} and {\color[HTML]{00B0F0} \textbf{blue}} fonts.} \label{table:CODQuanti}
            \vspace{1mm}
		\resizebox{2.07\columnwidth}{!}{
			\setlength{\tabcolsep}{1.4mm}
			\begin{tabular}{l|c|cccc|cccc|cccc|cccc} 
				\toprule
				\multicolumn{1}{c|}{}                                        & \multicolumn{1}{c|}{}                           & \multicolumn{4}{c|}{\textit{CHAMELEON} }                                                                                                                                         & \multicolumn{4}{c|}{\textit{CAMO} }                                                                                                                                             & \multicolumn{4}{c|}{\textit{COD10K} }                                                                                                                                          & \multicolumn{4}{c}{\textit{NC4K} }                                                                                                                        \\ \cline{3-18} 
				\multicolumn{1}{l|}{\multirow{-2}{*}{Methods}} & \multicolumn{1}{c|}{\multirow{-2}{*}{Backbones}} & {\cellcolor{gray!40}$M$~$\downarrow$}                                  & {\cellcolor{gray!40}$F_\beta$~$\uparrow$}                               & {\cellcolor{gray!40}$E_\phi$~$\uparrow$}                               & \multicolumn{1}{c|}{\cellcolor{gray!40}$S_\alpha$~$\uparrow$}                                   & {\cellcolor{gray!40}$M$~$\downarrow$}                                  & {\cellcolor{gray!40}$F_\beta$~$\uparrow$}                               & {\cellcolor{gray!40}$E_\phi$~$\uparrow$}                               & \multicolumn{1}{c|}{\cellcolor{gray!40}$S_\alpha$~$\uparrow$}                                   & {\cellcolor{gray!40}$M$~$\downarrow$}                                  & {\cellcolor{gray!40}$F_\beta$~$\uparrow$}                               & {\cellcolor{gray!40}$E_\phi$~$\uparrow$}                               & \multicolumn{1}{c|}{\cellcolor{gray!40}$S_\alpha$~$\uparrow$}                                   & {\cellcolor{gray!40}$M$~$\downarrow$}                                  & {\cellcolor{gray!40}$F_\beta$~$\uparrow$}                               & {\cellcolor{gray!40}$E_\phi$~$\uparrow$}                               & \multicolumn{1}{c}{\cellcolor{gray!40}$S_\alpha$~$\uparrow$}                                   \\ \midrule %\midrule
				\multicolumn{18}{c}{Common Setting: Single Input Scale and Single Stage}                                   \\ \midrule
				\multicolumn{1}{l|}{SINet~\cite{fan2020camouflaged}}& \multicolumn{1}{c|}{ResNet50}                   & 0.034                                 & 0.823                                 & 0.936                                 & \multicolumn{1}{c|}{0.872}                                 & 0.092                                 & 0.712                                 & 0.804                                 & \multicolumn{1}{c|}{0.745}                                 & 0.043                                 & 0.667                                 & 0.864                                 & \multicolumn{1}{c|}{0.776}                                 & 0.058                                 & 0.768                                 & 0.871                                 & 0.808                                 \\
				\multicolumn{1}{l|}{LSR~\cite{lv2021simultaneously}}                       & \multicolumn{1}{c|}{ResNet50}                   & 0.030                                 & 0.835                                 & 0.935                                 & \multicolumn{1}{c|}{0.890}                                 & 0.080                                 & 0.756                                 & 0.838                                 & \multicolumn{1}{c|}{0.787}                                 & 0.037                                 & 0.699                                 & 0.880                                 & \multicolumn{1}{c|}{0.804}                                 & 0.048 & 0.802 & 0.890                                 & 0.834                                 \\
                \multicolumn{1}{l|}{FEDER~\cite{He2023Camouflaged}} & \multicolumn{1}{c|}{ResNet50}  & {\color[HTML]{00B0F0}\textbf{0.028}} & {\color[HTML]{00B0F0}\textbf{0.850}} & 0.944 & \multicolumn{1}{c|}{0.892} & {{0.070}} & {0.775} & 0.870 & \multicolumn{1}{c|}{0.802} & 0.032 & 0.715 & 0.892 & \multicolumn{1}{c|}{0.810} & {{0.046}} & {{0.808}} & {{0.900}} & {{0.842}} \\
                \multicolumn{1}{l|}{FGANet~\cite{zhaiexploring}}                     & \multicolumn{1}{c|}{ResNet50}                   & 0.030                                 & 0.838                                 & {\color[HTML]{00B0F0}\textbf{0.945}}                                 & 0.891                                 & 0.070  & 0.769  & 0.865  & \multicolumn{1}{c|}{0.800}  & 0.032   & 0.708 & 0.894  & \multicolumn{1}{c|}{0.803}                                 & 0.047                                & 0.800                                 & 0.891                                 & 0.837                                 \\
                \multicolumn{1}{l|}{FocusDiff~\cite{zhao2025focusdiffuser}} & \multicolumn{1}{c|}{ResNet50} & {\color[HTML]{00B0F0}\textbf{0.028}} & 0.843 & 0.938 & 0.890  & {\color[HTML]{FF0000} \textbf{0.069}} & 0.772 & {\color[HTML]{FF0000} \textbf{0.883}} & {\color[HTML]{00B0F0} \textbf{0.812}} & {\color[HTML]{00B0F0} \textbf{0.031}} & {\color[HTML]{00B0F0} \textbf{0.730}} & {\color[HTML]{00B0F0} \textbf{0.897}} & 0.820 & {\color[HTML]{00B0F0} \textbf{0.044}} & {\color[HTML]{00B0F0} \textbf{0.810}} & {\color[HTML]{00B0F0} \textbf{0.902}} &{\color[HTML]{00B0F0} \textbf{0.850}}    \\
                
                \multicolumn{1}{l|}{FSEL~\cite{sun2025frequency}} & \multicolumn{1}{c|}{ResNet50} & 0.029 & 0.847 & 0.941 & {\color[HTML]{00B0F0} \textbf{0.893}}  & {\color[HTML]{FF0000} \textbf{0.069}} & {\color[HTML]{00B0F0}\textbf{0.779}} & {\color[HTML]{00B0F0} \textbf{0.881}} & {\color[HTML]{FF0000} \textbf{0.816}} & 0.032 & 0.722 & 0.891 & {\color[HTML]{00B0F0} \textbf{0.822}} & 0.045 & 0.807 & 0.901 & 0.847    \\ 
                
                \rowcolor{c2!20} \multicolumn{1}{l|}{RUN (Ours) }  & \multicolumn{1}{c|}{ResNet50} & {\color[HTML]{FF0000} \textbf{0.027}} & {\color[HTML]{FF0000} \textbf{0.855}} & {\color[HTML]{FF0000} \textbf{0.952}} & {\color[HTML]{FF0000} \textbf{0.895}} & 0.070 & {\color[HTML]{FF0000} \textbf{0.781}} & 0.868 & 0.806 & {\color[HTML]{FF0000} \textbf{0.030}} & {\color[HTML]{FF0000} \textbf{0.747}} & {\color[HTML]{FF0000} \textbf{0.903}} & {\color[HTML]{FF0000} \textbf{0.827}} & {\color[HTML]{FF0000} \textbf{0.042}} & {\color[HTML]{FF0000} \textbf{0.824}} & {\color[HTML]{FF0000} \textbf{0.908}} & {\color[HTML]{FF0000} \textbf{0.851}}     \\
                
                \midrule
				% \multicolumn{1}{l|}{SINet V2~\cite{fan2021concealed}}                  & \multicolumn{1}{c|}{Res2Net50}                  & 0.030                                  & 0.816                                 & 0.942                                 & \multicolumn{1}{c|}{0.888}                                 & 0.070                                  & 0.779                                 & 0.882                                 & \multicolumn{1}{c|}{0.822}                                 & 0.037                                 & 0.682                                 & 0.887                                 & \multicolumn{1}{c|}{0.815}                                 & 0.048                                 & 0.792                                 & 0.903                                 & 0.847                                 \\
				\multicolumn{1}{l|}{BSA-Net~\cite{zhu2022can}}            & \multicolumn{1}{c|}{Res2Net50}                  & 0.027                                 & 0.851                                 & 0.946                                 & \multicolumn{1}{c|}{0.895}                                 & 0.079                                 & 0.768                                 & 0.851                                 & \multicolumn{1}{c|}{0.796}                                 & 0.034                                 & 0.723                                 & 0.891                                 & \multicolumn{1}{c|}{0.818}                                 & 0.048                                 & 0.805                                 & 0.897                                 & 0.841                                 \\
				% \multicolumn{1}{l|}{BGNet~\cite{sun2022boundary}}                   & \multicolumn{1}{c|}{Res2Net50}                   & 0.029                                 & 0.835                                 & 0.944                                 & \multicolumn{1}{c|}{0.895}                                 & 0.073                                 & 0.744                                 & 0.870                                 & \multicolumn{1}{c|}{0.812}                                 & 0.033                                 & 0.714                                 & 0.901                                 & \multicolumn{1}{c|}{0.831}                                 & 0.044                                 & 0.786                                 & 0.907                                 & 0.851                                 \\
                \multicolumn{1}{l|}{FEDER~\cite{He2023Camouflaged} }                       & \multicolumn{1}{c|}{Res2Net50}   & 0.026 & 0.856  & 0.947  & 0.903  & \textbf{0.066}  & 0.807  & 0.897  & 0.836  & 0.029  & 0.748  & 0.911  & 0.844  & 0.042  & 0.824  & 0.913  & \textbf{0.862}               \\  
				% \multicolumn{1}{l|}{ICEG~\cite{he2023strategic} }                       & \multicolumn{1}{c|}{Res2Net50}                  & \textbf{0.025}& \textbf{0.869} & \textbf{0.958}& \textbf{0.908} & \textbf{0.066}  & \textbf{0.808} & \textbf{0.903} & \textbf{0.838} & \textbf{0.028}  & \textbf{0.752} & \textbf{0.914} & \textbf{0.845}   & \textbf{0.042} & \textbf{0.828}  & \textbf{0.917} & \textbf{0.867} \\  
                \rowcolor{c2!20} \multicolumn{1}{l|}{RUN (Ours) }  & \multicolumn{1}{c|}{Res2Net50} & \textbf{0.024} & \textbf{0.879} & \textbf{0.956} & \textbf{0.907} & \textbf{0.066} & \textbf{0.815} & \textbf{0.905} & \textbf{0.843}  & \textbf{0.028} & \textbf{0.764} & \textbf{0.914} & \textbf{0.849} & \textbf{0.041} & \textbf{0.830} & \textbf{0.917} &0.859                \\
                \midrule
				\multicolumn{1}{l|}{HitNet~\cite{hu2022high}}                & \multicolumn{1}{c|}{PVT V2}                  & 0.024                                 & 0.861                                 & 0.944                                 & \multicolumn{1}{c|}{0.907}                                 & 0.060                                 & 0.791                                 & 0.892                                 & \multicolumn{1}{c|}{0.834}                                 & 0.027                                 & 0.790                                 & 0.922                                 & \multicolumn{1}{c|}{0.847}                                 & 0.042                                 & 0.825                                 & 0.911                                 & 0.858                                 \\
                % \multicolumn{1}{l|}{OAFormer~\cite{yang2023oaformer} }               & \multicolumn{1}{c|}{PVT V2} & 0.023 & 0.868 & 0.961 & 0.904 & 0.048 & 0.849 & 0.924 & 0.867 & 0.025 &0.795 & 0.927 & 0.860 & 0.032 & 0.857 & 0.934 & 0.883 \\
				% \multicolumn{1}{l|}{ICEG~\cite{he2023strategic} }               & \multicolumn{1}{c|}{PVT V2}                  & \textbf{0.022}                                 & 0.879 & 0.957  & 0.913 & 0.043 & 0.863 & 0.933  & \multicolumn{1}{c|}{\textbf{0.876}}  &0.022 & 0.805  & 0.938 & 0.871  & 0.030 &0.869  & 0.941 & 0.890 \\ 
                \multicolumn{1}{l|}{CamoFocus~\cite{khan2024camofocus} }               & \multicolumn{1}{c|}{PVT V2} & 0.023 & 0.869 & 0.953 & 0.906 & \textbf{0.044} & \textbf{0.861} & 0.924 &0.870 & 0.022 & \textbf{0.818} & 0.931 & 0.868 & 0.031 & 0.862 &0.932 &0.886 \\
                \rowcolor{c2!20} \multicolumn{1}{l|}{RUN (Ours)} &  \multicolumn{1}{c|}{PVT V2} & \textbf{0.021} & \textbf{0.877} & \textbf{0.958} & \textbf{0.916} & 0.045 &  \textbf{0.861} &  \textbf{0.934} & \textbf{0.877} & \textbf{0.021} &  0.810 & \textbf{0.941} &  \textbf{0.878} & \textbf{0.030} & \textbf{0.868} & \textbf{0.940} & \textbf{0.892} \\ \midrule
				\multicolumn{18}{c}{Other Setting: Multiple Input Scales (MIS)} \\ 
				 \midrule
				\multicolumn{1}{l|}{ZoomNet~\cite{pang2022zoom}}            & \multicolumn{1}{c|}{ResNet50}                   & 0.024                                 & 0.858                                 & 0.943                                 & \multicolumn{1}{c|}{0.902}                                 & 0.066                                 & 0.792                                 & 0.877                                 & \multicolumn{1}{c|}{0.820}                                 & 0.029                                 & 0.740                                 & 0.888                                 & \multicolumn{1}{c|}{0.838}                        & 0.043                                 & 0.814                                 & 0.896                                 & 0.853                                 \\
                 % \multicolumn{1}{l|}{ZoomNext~\cite{pang2024zoomnext}} & \multicolumn{1}{c|}{ResNet50}  & 0.021 & 0.868 & 0.956 & 0.908 & 0.065 & 0.798 & 0.888 & 0.833 & 0.026 & 0.768 & 0.915 & 0.861 & 0.037 & 0.833 & 0.914 &0.874     \\
                \multicolumn{1}{l|}{FEDER~\cite{He2023Camouflaged}} & \multicolumn{1}{c|}{ResNet50}  & 0.023 & 0.869 & 0.959 & 0.906 & \textbf{0.064} & 0.801 & 0.893 & 0.827 & 0.028 & 0.756 & 0.913 & 0.837 & 0.041 & 0.832 & 0.915 & 0.859   \\ 
                % \multicolumn{1}{l|}{FSEL~\cite{sun2025frequency}} & \multicolumn{1}{c|}{ResNet50}  & --- & --- & --- & --- & 0.068 & 0.799 & 0.890 & 0.826 & 0.029 & 0.754 & 0.903 & 0.839 & \textbf{0.040} & 0.828 & 0.917 & 0.861 \\
                \rowcolor{c2!20} \multicolumn{1}{l|}{RUN (Ours)} & \multicolumn{1}{c|}{ResNet50}  & \textbf{0.022} & \textbf{0.878} & \textbf{0.967} & \textbf{0.911} & \textbf{0.064} & \textbf{0.807} & \textbf{0.902} & \textbf{0.832} & \textbf{0.027} & \textbf{0.772} & \textbf{0.920} &\textbf{0.843} & \textbf{0.040} & \textbf{0.836} & \textbf{0.922} & \textbf{0.868}\\

                \midrule
				% \multicolumn{1}{l|}{OCEG (Ours)}          & \multicolumn{1}{c|}{ResNet50} & \textbf{0.023} &\textbf{0.869} &\textbf{0.960} &\textbf{0.907} &\textbf{0.063} &\textbf{0.805} &\textbf{0.891} &0.832 &\textbf{0.028} & \textbf{0.756} &\textbf{0.915} &\textbf{0.842} & \textbf{0.041} & \textbf{0.833} &\textbf{0.914} &0.870\\ \midrule
				\multicolumn{18}{c}{Other Setting: Multiple Stages (MS)}  \\ \midrule
				\multicolumn{1}{l|}{SegMaR-4~\cite{jia2022segment}}   & \multicolumn{1}{c|}{ResNet50}                   & 0.025                        & 0.855                                 & 0.955                                 & \multicolumn{1}{c|}{0.906}                                 & 0.071                                 & 0.779                                 & 0.865                                 & \multicolumn{1}{c|}{0.815}                                 & 0.033                                 & 0.737                                 & 0.896                                 & \multicolumn{1}{c|}{0.833}                                 & 0.047                                 & 0.793                                 & 0.892                                 & 0.845                                 \\
                \multicolumn{1}{l|}{FEDER-4~\cite{He2023Camouflaged}}         & \multicolumn{1}{c|}{ResNet50}                  & 0.025 & 0.874  & 0.964  & 0.907  & 0.067  & 0.809  & 0.886  & 0.822  & 0.028  & 0.752  & 0.917  & 0.851  & 0.042  & 0.827  & 0.917  & 0.863         \\ 
				% \multicolumn{1}{l|}{ICEG-4~\cite{he2023strategic}}         & \multicolumn{1}{c|}{ResNet50}                   & \textbf{0.024}                        & 0.870                        & 0.961                        & 0.907                        & 0.067                        & 0.802                        & 0.884                        & 0.823                        & 0.028                        & 0.755                        & 0.920                        &0.843                       & 0.043                        & 0.824                        & 0.915                        & 0.860         \\ 
                \rowcolor{c2!20}\multicolumn{1}{l|}{RUN-4 (Ours) } & \multicolumn{1}{c|}{ResNet50} & \textbf{0.024} &\textbf{0.889} &\textbf{0.968}&\textbf{0.913}& \textbf{0.066} &\textbf{0.815} &\textbf{0.893} &\textbf{0.829} &\textbf{0.027} &\textbf{0.769} &\textbf{0.926} &\textbf{0.857} &\textbf{0.041} &\textbf{0.833}&\textbf{0.925}&\textbf{0.870} \\ \bottomrule            
		\end{tabular}}
		\vspace{-7mm}
	\end{table*}
\begin{table}[ht]
\centering
\caption{Results on polyp image segmentation.
        % The best two results are in {\color[HTML]{FF0000} \textbf{red}} and {\color[HTML]{00B0F0} \textbf{blue}} fonts.
        } \label{table:PISQuanti}
	\resizebox{1\columnwidth}{!}{
		\setlength{\tabcolsep}{0.8mm}
	\begin{tabular}{l|ccc|ccc}
		\toprule 
		\multirow{2}{*}{Methods} & \multicolumn{3}{c|}{\textit{CVC-ColonDB} }  & \multicolumn{3}{c}{\textit{ETIS} } 
        \\ \cline{2-7} 
		& \multicolumn{1}{c}{\cellcolor{gray!40}mDice~$\uparrow$} & \multicolumn{1}{c}{\cellcolor{gray!40}mIoU~$\uparrow$} & \multicolumn{1}{c|}{\cellcolor{gray!40}$S_\alpha$~$\uparrow$} & \multicolumn{1}{c}{\cellcolor{gray!40}mDice~$\uparrow$} & \multicolumn{1}{c}{\cellcolor{gray!40}mIoU~$\uparrow$} & \multicolumn{1}{c}{\cellcolor{gray!40}$S_\alpha$~$\uparrow$} \\ \midrule
		% U-Net~\cite{ronneberger2015u}    & 0.512   & 0.444 & 0.712  &  0.398 & 0.335 & 0.684\\
		PraNet~\cite{fan2020pranet} & 0.709 & 0.640  & 0.819  &  0.628 & 0.567 & 0.794 \\
		% MSNet~\cite{zhao2021automatic} & 0.755 & 0.678 & 0.836  & 0.719 & 0.664 & 0.840 \\
		% TGANet~\cite{tomar2022tganet}    & 0.722 & 0.661 & 0.823 & 0.706 & 0.651 & 0.843  \\
		% LDNet~\cite{zhang2022lesion} & {0.764} & 0.683  & 0.834 & 0.744 & 0.683 & 0.839   \\
        CASCADE~\cite{rahman2023medical} & 0.809 & 0.731 & 0.867 & 0.781 & {\color[HTML]{00B0F0} \textbf{0.706}} & 0.853  \\
        PolypPVT~\cite{dong2023polyp}  & 0.808 &0.727 & 0.865 & {\color[HTML]{00B0F0} \textbf{0.787}} & {\color[HTML]{00B0F0} \textbf{0.706}} & {\color[HTML]{00B0F0} \textbf{0.871}} \\
        CoInNet~\cite{jain2023coinnet} & 0.797 &0.729 &{\color[HTML]{00B0F0} \textbf{0.875}} & 0.759 & 0.690 & 0.859 \\
        LSSNet~\cite{wang2024lssnet} & {\color[HTML]{00B0F0} \textbf{0.820}} & {\color[HTML]{00B0F0} \textbf{0.741}} & 0.867 & 0.779 & 0.701 & 0.867 \\
        \rowcolor{c2!20} RUN (Ours) & {\color[HTML]{FF0000} \textbf{0.822}} & {\color[HTML]{FF0000} \textbf{0.742}} & {\color[HTML]{FF0000} \textbf{0.880}} & {\color[HTML]{FF0000} \textbf{0.788}} &{\color[HTML]{FF0000} \textbf{0.709}} & {\color[HTML]{FF0000} \textbf{0.878}}   \\
	 \bottomrule                      
	\end{tabular}}\label{table:polyp}
		\vspace{-0.7cm}
	\end{table} 

\begin{table}[ht]
	\centering
        \caption{Results on medical tubular object segmentation.
        % The best two results are in {\color[HTML]{FF0000} \textbf{red}} and {\color[HTML]{00B0F0} \textbf{blue}} fonts.
        } \label{table:MTOSQuanti}
	\resizebox{1\columnwidth}{!}{
		\setlength{\tabcolsep}{0.8mm}
	\begin{tabular}{l|ccc|ccc}
		\toprule 
		\multirow{2}{*}{Methods} & \multicolumn{3}{c|}{\textit{DRIVE} }  & \multicolumn{3}{c}{\textit{CORN} } 
        \\ \cline{2-7} 
		& \multicolumn{1}{c}{\cellcolor{gray!40}mDice~$\uparrow$} & {\cellcolor{gray!40}AUC~$\uparrow$}                               & \multicolumn{1}{c|}{\cellcolor{gray!40}SEN~$\uparrow$} & \multicolumn{1}{c}{\cellcolor{gray!40}mDice~$\uparrow$} & {\cellcolor{gray!40}AUC~$\uparrow$}                               & \multicolumn{1}{c|}{\cellcolor{gray!40}SEN~$\uparrow$}  \\ \midrule
		% U-Net~\cite{ronneberger2015u} & 0.782                     & 0.978                   & 0.780                    & 0.450                     & 0.855                   & 0.507   \\
		% CS-Net~\cite{mou2019cs} & 0.775                     & 0.980                   & {\color[HTML]{00B0F0} \textbf{0.835}}                    & 0.582                     & 0.951                   & 0.773     \\
		% DeepVessel~\cite{tetteh2020deepvesselnet} & 0.783                     & 0.979                   & 0.760                    & 0.613                     & 0.946                   & 0.782       \\
		% TransUNet~\cite{chen2021transunet}  & 0.781                     & 0.938                   & 0.817                    & 0.620                     & 0.953                   & 0.775    \\
		CS2-Net~\cite{mou2021cs2}  & 0.795                     & {\color[HTML]{00B0F0} \textbf{0.983}}                   & 0.822                    & 0.607                     & 0.960                   & 0.817       \\
        % GT-DLA~\cite{li2022global} & & &  & --- & --- & ---   \\
        DSCNet~\cite{qi2023dynamic}   & 0.805                     & 0.955                   & 0.830                    & 0.618                     & {\color[HTML]{FF0000} \textbf{0.964}}                   & 0.856     \\
        SGAT~\cite{lin2023stimulus}   & {\color[HTML]{00B0F0} \textbf{0.806}}                     & 0.953                   & {\color[HTML]{00B0F0} \textbf{0.832}}                    & 0.639                     & 0.961                   & 0.853      \\
        TAU~\cite{gupta2024topology}    & 0.798                     & 0.977                   & 0.825                    & 0.643                     & 0.949                   & {\color[HTML]{00B0F0} \textbf{0.859}}      \\
        FFM~\cite{huang2025representing}    & 0.791                     & 0.972                   & 0.830                    & {\color[HTML]{00B0F0} \textbf{0.647}}                     & 0.952                   & 0.835              \\
        \rowcolor{c2!20} RUN (Ours) & {\color[HTML]{FF0000} \textbf{0.812}}                     & {\color[HTML]{FF0000} \textbf{0.985}}                   & {\color[HTML]{FF0000} \textbf{0.845}}                    & {\color[HTML]{FF0000} \textbf{0.652}}                      & {\color[HTML]{00B0F0} \textbf{0.962}}                   & {\color[HTML]{FF0000} \textbf{0.870}}           \\
	 \bottomrule                      
	\end{tabular}}\label{table:vessel}
    \vspace{-0.7cm}
		\end{table}

\section{Experiments}
% \subsection{Experimental Setup}
\noindent \textbf{Implementation details}. We implement our method using PyTorch on two RTX4090 GPUs. In line with~\cite{fan2020camouflaged}, we incorporate deep features from encoder-shaped networks into our framework. All images are resized to $352 \times 352$ for the training and testing phases. During training, we use the Adam optimizer with momentum parameters $(0.9, 0.999)$. The batch size is set to 36, and the initial learning rate is configured to 0.0001, which is reduced by 0.1 every 80 epochs. The stage number $K$ is set as 4. Additional parameters inherited from traditional methods are optimized in a learnable manner with random initialization.

\subsection{Comparative Evaluation}
% We perform experiments on several challenging COS tasks and compare our performance with SOTA methods using commonly used metrics. For fairness, all segmentation results are evaluated using consistent evaluation tools specific to each task. Apart from COD, other tasks have a limited number of publicly available methods, which restricts the scope of quantitative analysis. Detailed information on the datasets and evaluation metrics is provided in~\cref{Sec:Dataset}. We also prove our superiority in salient object detection, which can be seen in~\cref{Sec:SOD}.
We conduct experiments on various COS tasks and compare our performance with SOTA methods using standard metrics. 
Details on datasets and metrics are in~\cref{Sec:Dataset}. Our superiority in concealed defect detection and salient object detection is verified in~\cref{Sec:CDD,Sec:SOD}. For fairness, all results are evaluated with consistent task-specific evaluation tools. Except for COD, other tasks have few publicly open-sourced methods, limiting quantitative analysis.

% \begin{figure*}[t]
% \setlength{\abovecaptionskip}{0cm}
% 	\centering
% 	\includegraphics[width=\linewidth]{Figure/MedQuali-try1.pdf}\vspace{-4mm}
% 	\caption{Visual comparison on PIS, MTOS, and TOD tasks.}
% 	\label{fig:MedQuali1}
% 	\vspace{-7mm}
% \end{figure*}

\noindent \textbf{Camouflaged object detection}. 
As shown in~\cref{table:CODQuanti}, our method achieves SOTA performance across all three settings. In the common setting, it outperforms competing approaches across all three backbones: ResNet50~\cite{he2016deep}, Res2Net50~\cite{gao2019res2net}, and PVT V2~\cite{wang2022pvt}. This superior performance on four datasets, particularly on the largest dataset, \textit{COD10K}, and the largest testing dataset, \textit{NC4K}, underscores the robustness and generalization capabilities of our RUN framework. 
Furthermore, in the MIS and MS settings, our RUN adheres to the evaluation protocols of FEDER~\cite{He2023Camouflaged} and delivers improved results over existing methods. 
% This further corroborates its superior performance. 
As illustrated in ~\cref{fig:MedQuali1}, our method generates more complete and accurate segmentation maps. This is attributed to our jointly reversible modeling at both the mask and RGB levels.
% , which effectively reduces false-positive and false-negative regions.

% This leading performance on four datasets, especially in the largest dataset \textit{COD10K} and the largest testing dataset \textit{NC4K}, comprehensively demonstrate the superiority of our RUN in robustness and generalization. Furthermore, in MIS and MS settings, our RUN strictly follows the practice of FEDER~\cite{He2023Camouflaged} and also achieves better results than existing methods. This further validates our advanced performance. As shown in Fig. , our method can generate more complete and accurate segmentation maps. This is attributed to our jointly reversible modeling at both mask level and RGB space, effectively eliminating false-positive and false-negative regions.

\begin{table}[tbp!]
	\setlength{\abovecaptionskip}{0cm} 
	\setlength{\belowcaptionskip}{-0.2cm}
	\centering
        \caption{Results on transparent object detection. 
        % Following the common practice, we employ ResNext101 as our backbone.
        } \label{table:TODQuanti}
	\resizebox{\columnwidth}{!}{
		\setlength{\tabcolsep}{1mm}
	\begin{tabular}{l|ccc|ccc}\toprule 

		\multicolumn{1}{l|}{}                          & \multicolumn{3}{c|}{\textit{GDD} } & \multicolumn{3}{c}{\textit{GSD} }  \\ \cline{2-7} 
		\multicolumn{1}{l|}{\multirow{-2}{*}{Methods}} & \cellcolor{gray!40}mIoU~$\uparrow$&\cellcolor{gray!40}$F_\beta^{max}$~$\uparrow$&\cellcolor{gray!40} $M$~$\downarrow$& \cellcolor{gray!40}mIoU~$\uparrow$&\cellcolor{gray!40}$F_\beta^{max}$~$\uparrow$&\cellcolor{gray!40} $M$~$\downarrow$ \\ \midrule
		% PMD~\cite{lin2020progressive}& 0.870                                 & 0.930                                 & 0.067                                                                  & 0.817                                 & 0.890                                 & 0.061  \\
		GDNet~\cite{mei2020don}                                          & 0.876                                 & 0.937                                 & 0.063                                                              & 0.790                                 & 0.869                                 & 0.069    \\
		% GateNet~\cite{zhao2020suppress}                                        & 0.817                                 & 0.931                                 & 0.073                                                            & 0.689                                 & 0.898                                 & 0.073  \\
		% GlassNet~\cite{lin2021rich}                                       & 0.881                                 & 0.932                                 & 0.059                                                      & 0.836                                 & 0.901                                 & 0.055  \\
		EBLNet~\cite{he2021enhanced} & 0.870                                 & 0.922                                 & 0.064                                  & 0.817                                 & 0.878                                 & 0.059 \\
        % PGSNet~\cite{yu2022progressive} &  0.878 & 0.901 & 0.062 & 0.836 & 0.868 & 0.054 \\
		% CSNet~\cite{cheng2021highly}                                          & 0.773                                 & 0.876                                 & 0.135                                 & 11.33                                & 0.666                                 & 0.805                                 & 0.135                                 & 14.76                                               \\
		% PGNet~\cite{xie2022pyramid}                                          & 0.857                                 & 0.930                                 & 0.074                                 & 6.82                                 & 0.805                                 & 0.897                                 & 0.068                                 & 7.88                                                \\
		% GDNet-B~\cite{mei2022large}                                        & 0.878                                 & 0.939                                 & 0.061         & 0.792                                 & 0.874                                 & 0.066  \\
        RFENet~\cite{fan2023rfenet} & 0.886 & 0.938 & 0.057 & {\color[HTML]{00B0F0} \textbf{0.865}} & {\color[HTML]{00B0F0} \textbf{0.931}} & {\color[HTML]{00B0F0} \textbf{0.048}} \\
        % GMB~\cite{qi2024glass} &  0.892 & 0.938 & {\color[HTML]{00B0F0} \textbf{0.054}}  & 0.832 & 0.895 & 0.051   \\
        IEBAF~\cite{han2023internal} & 0.887 & {\color[HTML]{00B0F0} \textbf{0.944}} & 0.056  &  0.861 & 0.926 & 0.049  \\
		% HCM~\cite{xiao2023concealed} & 0.908 & 0.946 & 0.045 & 4.42 \\
        GhostingNet~\cite{yan2024ghostingnet} & {\color[HTML]{00B0F0} \textbf{0.893}}  & 0.943 & {\color[HTML]{00B0F0} \textbf{0.054}} & 0.838 & 0.904 & 0.055   \\
	\rowcolor{c2!20} 	RUN (Ours)  & {\color[HTML]{FF0000} \textbf{0.895}} & {\color[HTML]{FF0000} \textbf{0.952}} & {\color[HTML]{FF0000} \textbf{0.051}} & {\color[HTML]{FF0000} \textbf{0.866}} & {\color[HTML]{FF0000} \textbf{0.938}} & {\color[HTML]{FF0000} \textbf{0.043}}   \\ \bottomrule  \end{tabular}}
\vspace{-9mm}
\end{table}
\noindent \textbf{Medical concealed object segmentation}. We conducted experiments on two medical COS tasks, including polyp image segmentation (\textit{CVC-ColonDB} and \textit{ETIS} datasets) and medical tubular object segmentation (\textit{DRIVE} and \textit{CORN} datasets).
% , fundus vessel segmentation (\textit{STARE}), and corneal nerve segmentation (\textit{CORN}). 
Considering that recent SOTAs commonly use Transformer-based encoders, we adopt PVT V2 as our default encoder. As shown in~\cref{table:PISQuanti,table:MTOSQuanti}, our method achieves top performance across three tasks. Furthermore, the results in~\cref{fig:MedQuali1} confirm the effect of our approach in segmenting small polyps and fine vessels and nerves. 
% This can be attributed to our RSS module, which employs a reversible strategy to explore concealed objects both locally and globally, enabling the extraction of subtle details.

\noindent \textbf{Transparent object detection}. Accurately segmenting transparent objects is crucial for autonomous driving.
% by enabling the detection of transparent obstacles. 
As demonstrated in~\cref{table:TODQuanti,fig:MedQuali1}, our RUN surpasses existing methods on two datasets, providing more precise segmentation of transparent objects compared to other approaches. These results highlight our potential to contribute to the advancement of autonomous driving.

\begin{figure*}[t]
\setlength{\abovecaptionskip}{0cm}
	\centering
	\includegraphics[width=\linewidth]{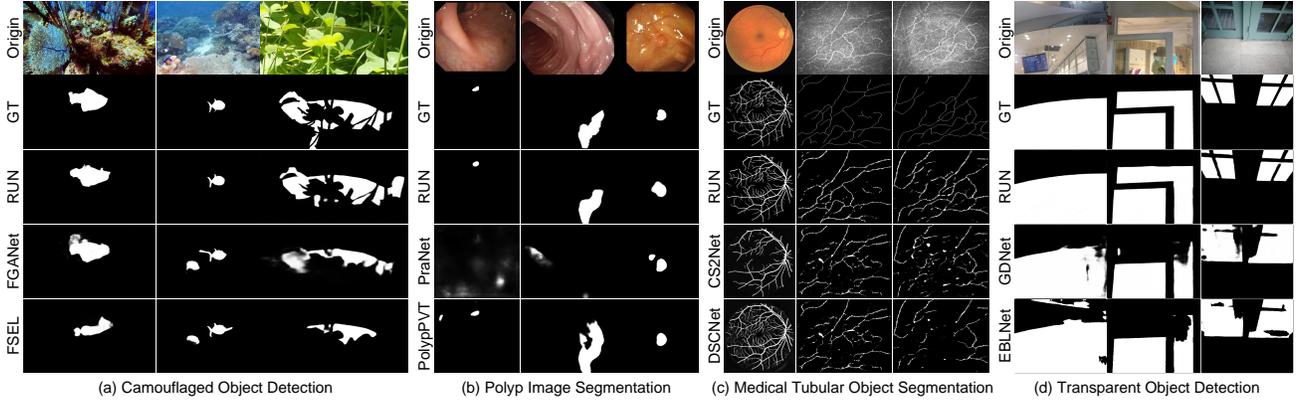}\vspace{-4mm}
	\caption{Visual comparison on COD, PIS, MTOS, and TOD tasks.}
	\label{fig:MedQuali1}
	\vspace{-7mm}
\end{figure*}
\begin{table*}[ht]
\centering
\begin{minipage}{1\textwidth}
\setlength{\abovecaptionskip}{0cm}
\caption{Ablation study in the COD task on \textit{COD10K}.}\label{Table:Ablation}
\resizebox{\columnwidth}{!}{ 
\setlength{\tabcolsep}{1.4mm}
\begin{tabular}{c|cccccc|ccc|c|c}
\toprule
\multirow{2}{*}{Metrics} & \multicolumn{6}{c|}{Effect of SOFS}                                             & \multicolumn{3}{c|}{Effect of ROBE}    &Fixed $\rightarrow$                       &\cellcolor{c2!20} {RUN } \\ \cline{2-10}
                         & $\mathbf{C}\rightarrow E(\mathbf{C})$ & w/o RSS & w/o VSS & w/o prior $\hat{\mathbf{M}}_k$  & w/o prior $\mathbf{B}_{k-1}$  &\multicolumn{1}{c|}{w/o $\mathbf{E}_k$} & $\mathcal{B}_1(\bigcdot) \rightarrow \mathcal{B}(\bigcdot)$ & $\mathcal{B}_2(\bigcdot) \rightarrow \mathcal{B}(\bigcdot)$ & \multicolumn{1}{c|}{w/o $\hat{\mathbf{C}}_k$} &  Learnable &    \cellcolor{c2!20} (Ours)                  \\  \midrule
$M$~$\downarrow$     & 0.053                & 0.034   & 0.031   & 0.032       & 0.031      & 0.031    & 0.030                 & 0.030                 & 0.031 & 0.032       &\cellcolor{c2!20} 0.030                       \\
$F_\beta$~$\uparrow$         & 0.617                & 0.710   & 0.740   & 0.728       & 0.735      & 0.733    & 0.746                 & 0.749                 & 0.736   &  0.731    &\cellcolor{c2!20} 0.747                       \\
$E_\phi$~$\uparrow$     & 0.825                & 0.887   & 0.897   & 0.891       & 0.893      & 0.890    & 0.905                 & 0.906                 & 0.898  &     0.896   &\cellcolor{c2!20} 0.903                       \\
$S_\alpha$~$\uparrow$     & 0.746                & 0.805   & 0.825   & 0.820       & 0.826      & 0.823    & 0.826                 & 0.828                 & 0.824  &   0.822   &\cellcolor{c2!20} 0.827       \\ \bottomrule      
\end{tabular}} 
\end{minipage}
\\ \vspace{-2mm}
	\begin{minipage}{.605\textwidth}
		\centering
		\setlength{\abovecaptionskip}{0cm}
		\caption{Effect of our COS model. CM, PM, OS, and DL are shorts for conventional model, proposed model, optimization solution, and deep learning.
		}
		\resizebox{\columnwidth}{!}{
			\setlength{\tabcolsep}{0.8mm}
			\begin{tabular}{c|ccccccc}
            \toprule
Metrics & PM+OS & CM1+DL & CM2+DL & CM3+DL & CM4+DL & CM5+DL & \cellcolor{c2!20}PM+DL (Ours) \\ \midrule
$M$~$\downarrow$   & 0.062 & 0.032  & 0.031  & 0.031  & 0.031  & 0.032  & \cellcolor{c2!20}0.030        \\
$F_\beta$~$\uparrow$ & 0.573 & 0.729  & 0.735  & 0.733  & 0.740  & 0.735  & \cellcolor{c2!20}0.747        \\
$E_\phi$~$\uparrow$   & 0.802 & 0.899  & 0.896  & 0.895  & 0.898  & 0.892  & \cellcolor{c2!20}0.903        \\
$S_\alpha$~$\uparrow$   & 0.733 & 0.823  & 0.824  & 0.821  & 0.823  & 0.823  & \cellcolor{c2!20}0.827  \\ \bottomrule
		\end{tabular}}\label{table:ablationModel}
		% \vspace{-0.1cm}
	\end{minipage} 
    \begin{minipage}{.386\textwidth}
		    \centering
		\setlength{\abovecaptionskip}{0cm}
		\caption{Analysis of stage number $K$. We have surpassed most compared methods when $K=2$.
		}
		\resizebox{\columnwidth}{!}{
			\setlength{\tabcolsep}{0.8mm}
			\begin{tabular}{c|ccccc} \toprule
Metrics & $K=1$     & $K=2$    & \cellcolor{c2!20}$K=4$ (Ours) & $K=6$ & $K=8$   \\  \midrule
$M$~$\downarrow$    & 0.033 & 0.031 &\cellcolor{c2!20} 0.030    & 0.030 &0.030 \\
$F_\beta$~$\uparrow$  & 0.715 & 0.727 &\cellcolor{c2!20} 0.747    & 0.749 & 0.751 \\
$E_\phi$~$\uparrow$   & 0.885 & 0.893 &\cellcolor{c2!20} 0.903    & 0.905 & 0.905 \\
$S_\alpha$~$\uparrow$   & 0.803 & 0.812 &\cellcolor{c2!20} 0.827    & 0.826 & 0.830 \\ \bottomrule
		\end{tabular}}\label{table:ablationStageNum}
		\end{minipage}
        \vspace{-6mm}
\end{table*}
\begin{table*}[ht]
	\centering
	\setlength{\abovecaptionskip}{0.1cm}
	% \caption{Results on extreme-challenging images on \textit{COD10K} in WSCOS.}\label{table:ExtremeImg} \vspace{-1mm}
	\begin{minipage}{.475\textwidth}
		\centering
		\setlength{\abovecaptionskip}{0cm}
		\caption{Results on small object images (1,084 images).
		}
		\resizebox{\columnwidth}{!}{
			\setlength{\tabcolsep}{1mm}
			\begin{tabular}{c|cccccc}
				\toprule
				Metrics & SegMaR & FEDER  & FGANet  & FocusDiff  &  FSEL& \cellcolor{c2!20} RUN (Ours)          \\ \midrule
				$M$~$\downarrow$ &0.049   & 0.044 & 0.044  & 0.042     & 0.043 &\cellcolor{c2!20} \textbf{0.040} \\
				$F_\beta$~$\uparrow$ &0.605 & 0.646 & 0.642  & 0.670     & 0.668 &\cellcolor{c2!20} \textbf{0.682} \\
				$E_\phi$~$\uparrow$ &0.831  & 0.855 & 0.852  & 0.859     & 0.847 &\cellcolor{c2!20} \textbf{0.866} \\
				$S_\alpha$~$\uparrow$   &0.765    & 0.777 & 0.776  & 0.781     & 0.776 &\cellcolor{c2!20} \textbf{0.789} \\ \bottomrule
		\end{tabular}}\label{table:small-object}
		% \vspace{-0.1cm}
	\end{minipage} \hspace{5mm}
    \begin{minipage}{.475\textwidth}
		    \centering
		\setlength{\abovecaptionskip}{0cm}
		\caption{Results on multi-object images (186 images).
		}
		\resizebox{\columnwidth}{!}{
			\setlength{\tabcolsep}{1mm}
			\begin{tabular}{c|cccccc}
				\toprule
				Metrics & SegMaR & FEDER  & FGANet  &  FocusDiff  &  FSEL& \cellcolor{c2!20} RUN (Ours)          \\ \midrule
				$M$~$\downarrow$  & 0.076  & 0.068 & 0.065  & 0.062     & 0.062 &\cellcolor{c2!20} \textbf{0.060} \\
				$F_\beta$~$\uparrow$  & 0.436 & 0.480 & 0.481  & 0.500     & 0.496 &\cellcolor{c2!20} \textbf{0.505} \\
				$E_\phi$~$\uparrow$  & 0.797 & 0.813 & 0.810  & 0.818     & 0.820 &\cellcolor{c2!20} \textbf{0.827} \\
				$S_\alpha$~$\uparrow$   & 0.695    & 0.709 & 0.709  & 0.716     & 0.717 &\cellcolor{c2!20} \textbf{0.730} \\ \bottomrule
		\end{tabular}}\label{table:Multi-object}
		\end{minipage}
\\ \vspace{1mm}
\begin{minipage}{0.485\textwidth}    
	\begin{minipage}{0.49\textwidth}
		\centering
		\includegraphics[width=\textwidth]{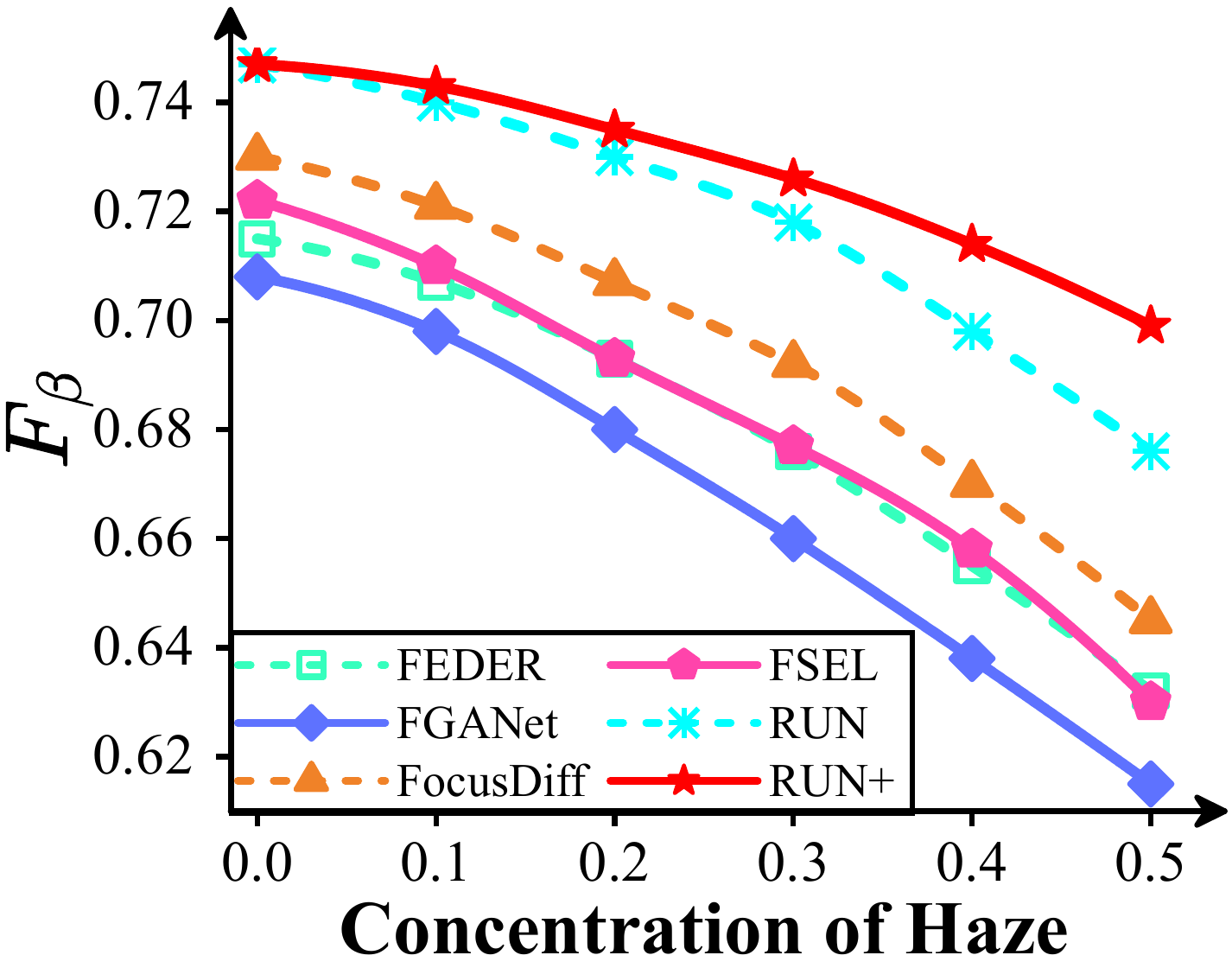}
		% \caption{Low-light image enhancement.}
	\end{minipage}
	\hfill
	\begin{minipage}{0.49\textwidth}  
		\centering 
		\includegraphics[width=\textwidth]{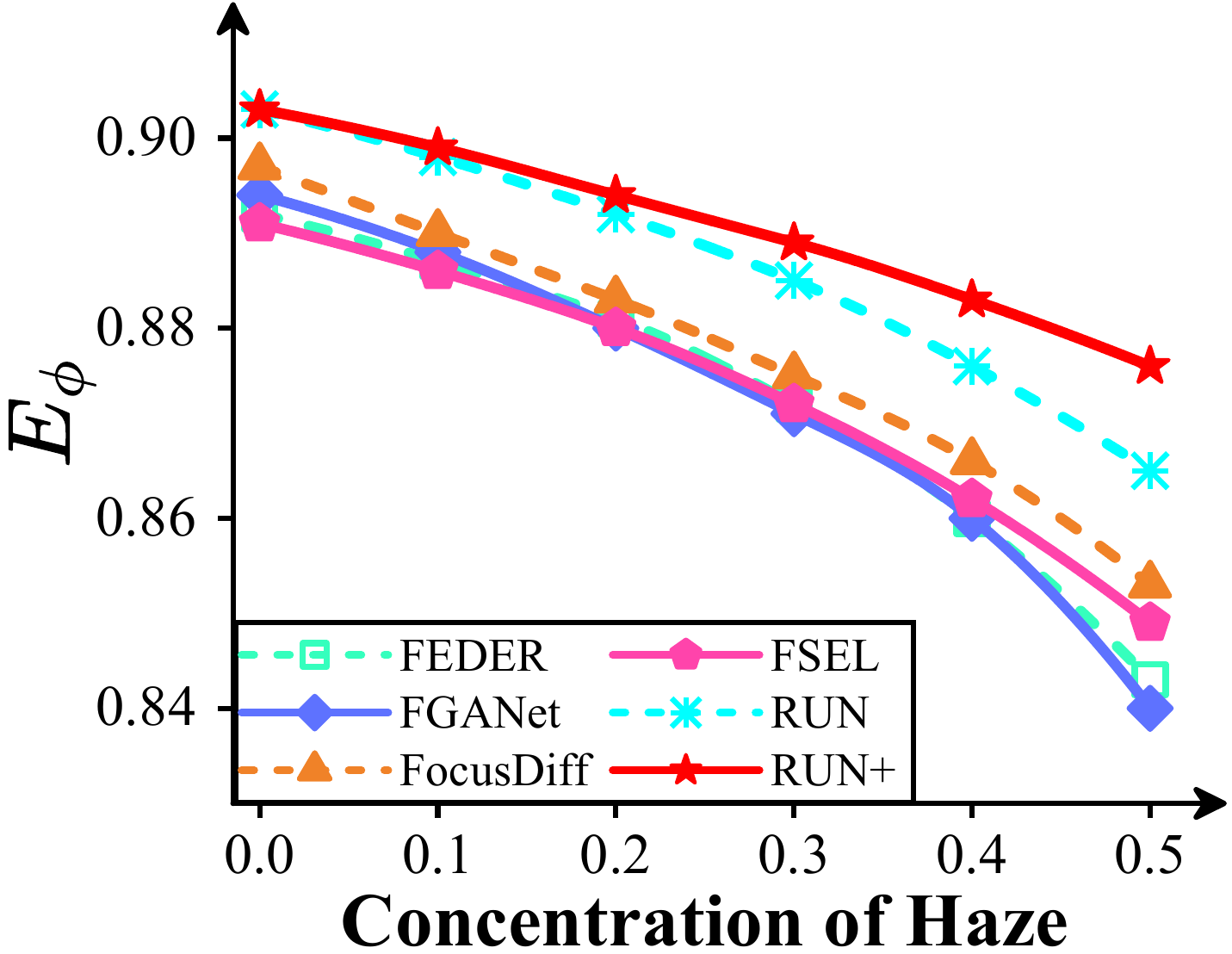}
		% \caption{Underwater image enhancement.}
	\end{minipage}
\vspace{-3mm}
	\captionof{figure}{Performance in degraded COS scenarios.
 }
	\label{fig:ExperDegradation}
\end{minipage} \hspace{5mm}
\begin{minipage}{0.475\textwidth}  
    \centering
	\includegraphics[width=\linewidth]{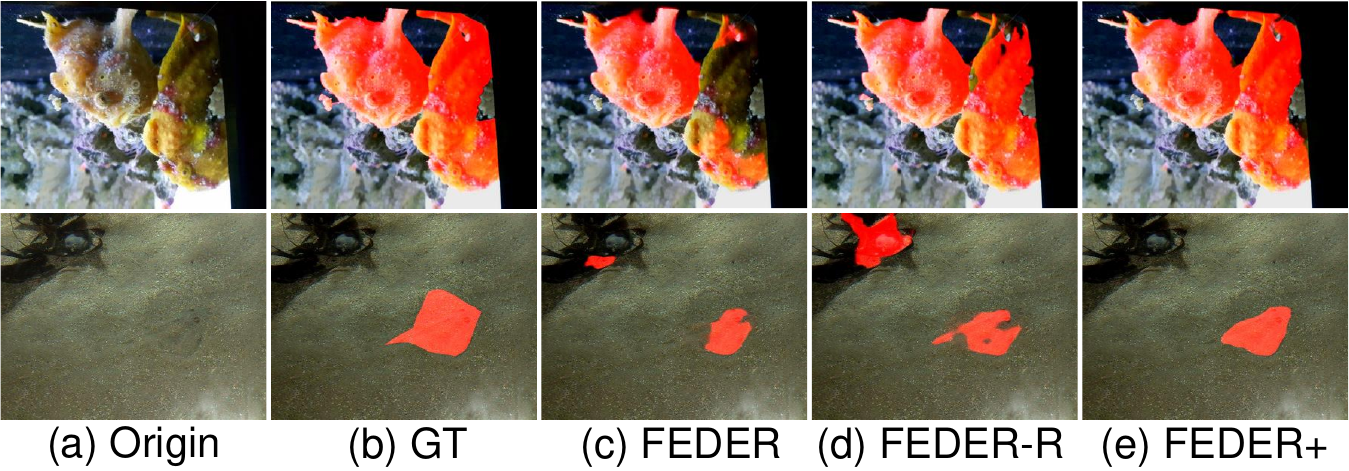}\vspace{-4mm}
	\captionof{figure}{Potential applications of RUN. The concealed object masks are highlighted in {\color{red}{red}} and overlaid on the original images.}
	\label{fig:Application}
\end{minipage}
\\ \vspace{-1mm}
	\begin{minipage}{.477\textwidth}
		    \centering
		\setlength{\abovecaptionskip}{0cm}
		\caption{Potential of RUN to serve as a refiner, where ``FEDER-R'' means refining FEDER's results with RUN.
		}
		\resizebox{\columnwidth}{!}{
			\setlength{\tabcolsep}{1mm}
			\begin{tabular}{c|cc|cc|cc}
				\toprule
				Metrics & FEDER &\cellcolor{c2!20} FEDER-R & FGANet &\cellcolor{c2!20} FGANet-R & FSEL  &\cellcolor{c2!20} FSEL-R \\ \midrule 
				$M$~$\downarrow$  & 0.032 &\cellcolor{c2!20} 0.031   & 0.032  &\cellcolor{c2!20} 0.032    & 0.032 &\cellcolor{c2!20} 0.031 \\
				$F_\beta$~$\uparrow$  & 0.715 &\cellcolor{c2!20} 0.721   & 0.708  &\cellcolor{c2!20} 0.716    & 0.722 &\cellcolor{c2!20} 0.725 \\
				$E_\phi$~$\uparrow$  & 0.892 &\cellcolor{c2!20} 0.897   & 0.894  &\cellcolor{c2!20} 0.897    & 0.891 &\cellcolor{c2!20} 0.890 \\
				$S_\alpha$~$\uparrow$   & 0.810 &\cellcolor{c2!20} 0.812   & 0.803  &\cellcolor{c2!20} 0.805    & 0.822 &\cellcolor{c2!20} 0.825 \\ \bottomrule
		\end{tabular}}\label{table:RefinerRUN}
		\end{minipage} \hspace{6mm}
    \begin{minipage}{.468\textwidth}
		\centering
		\setlength{\abovecaptionskip}{0cm}
		\caption{Generalization of RUN, where ``FEDER+'' means integrating our framework into FEDER.
		}
		\resizebox{\columnwidth}{!}{
			\setlength{\tabcolsep}{1mm}
			\begin{tabular}{c|cc|cc|cc}
				\toprule
				Metrics & FEDER &\cellcolor{c2!20} FEDER+ & FGANet &\cellcolor{c2!20} FGANet+ & FSEL  &\cellcolor{c2!20} FSEL+ \\ \midrule
				$M$~$\downarrow$ &0.032 &\cellcolor{c2!20} 0.031  & 0.032  &\cellcolor{c2!20}  0.031   & 0.032 &\cellcolor{c2!20} 0.030 \\
				$F_\beta$~$\uparrow$ &0.715 &\cellcolor{c2!20} 0.726  & 0.708  &\cellcolor{c2!20} 0.730   & 0.722 &\cellcolor{c2!20} 0.738 \\
				$E_\phi$~$\uparrow$ &0.892 &\cellcolor{c2!20} 0.902  & 0.894  &\cellcolor{c2!20} 0.901   & 0.891 &\cellcolor{c2!20} 0.905 \\
				$S_\alpha$~$\uparrow$   &0.810 &\cellcolor{c2!20} 0.816  & 0.803  &\cellcolor{c2!20} 0.808   & 0.822 &\cellcolor{c2!20} 0.830 \\ \bottomrule
		\end{tabular}}\label{table:GeneraRUN}
		% \vspace{-0.1cm}
	\end{minipage} 
    \vspace{-7mm}
\end{table*}	

\subsection{Ablation Study}
We conduct ablation studies on \textit{COD10K} of the COD task.

\noindent \textbf{Effect of SOFS}. As presented in~\cref{Table:Ablation}, replacing the deep features $E(\mathbf{C})$ with the concealed image results in performance decline, highlighting the critical role of incorporating deep features into the DUN-based framework. Additionally, we evaluate the impact of the state space-based structure by removing the RSS and VSS modules. The effectiveness of our reversible strategy is further validated by excluding the foreground prior $\hat{\mathbf{M}}_k$ and the background prior $\mathbf{B}_{k-1}$. Finally, we confirm the utility of integrating the auxiliary edge output, contributing to performance improvements.

\noindent \textbf{Effect of ROBE}. 
% We first prove that a simple reconstruction network is enough for 
As shown in~\cref{Table:Ablation}, when replacing $\mathcal{B}(\bigcdot)$ with other large-scale networks, \textit{i.e.}, the CNN-based network $\mathcal{B}_1(\bigcdot)$~\cite{he2023HQG} and Transformer-based network $\mathcal{B}_2(\bigcdot)$~\cite{he2023reti}, we observe no significant performance gains. This suggests that a simple network is sufficient for background extraction and image reconstruction. Furthermore, when the reconstructed output $\hat{\mathbf{C}}_k$ is removed, our RUN also produces suboptimal results.

\noindent \textbf{Other configurations in RUN}. We validate the effect of various configurations in RUN. As shown in~\cref{Table:Ablation}, allowing originally fixed parameters to be learnable enhances performance. Furthermore, we compare our model with CM1 to CM5, as detailed in~\cref{table:ablationModel}. CM1 corresponds to \cref{Eq:BasicModel1}. CM2 also employs \cref{Eq:FinalModel} for optimization but applies the $\ell_1$-norm to the first term and employs soft thresholding~\cite{he2023degradation} to solve $\mathcal{S}(\bigcdot)$. 
CM3-CM5 represent ablated versions of the refined mask and weighted map in \cref{Eq:FinalModel}: CM3 removes the weighted map $\mathbf{w}$. CM4 modifies the refined range of pixel values from $[0.1,0.4) \& (0.6,0.9]$ to $[0.1,0.3) \& (0.7,0.9]$, with corresponding adjustments to the weighted map. CM5 extends the range to include all pixel values, assigning 0.5 to the foreground region. 
As shown in~\cref{table:ablationModel}, our approach achieves superior performance across traditional solutions and learning-based unfolding strategies. 
Moreover, as verified in~\cref{table:ablationStageNum}, we analyze the optimal stage number for our method. To balance performance and computational efficiency, we set $K=4$. Under this configuration, the feature maps from the last four layers of the encoder are progressively sent to the four stages, with features from deeper layers transferred first.

% Here, we verify the effectiveness of other configurations in RUN. As illustrated in~\cref{Table:Ablation}, making originally fixed parameters to be learnable can greatly improve the segmentation performance. Additionally, as presented in~\cref{table:ablationModel}, we compared our model with CM1 (\cref{Eq:BasicModel1}) and CM2, where CM2 also selects \cref{Eq:FinalModel} as the object function but employs $\ell_1$ as the restriction of $\mathcal{S}(\bigcdot)$ and use soft thresholding to solve this, like what is conducted in~\cite{he2023degradation}. Our method achieves the best performance in both the traditional optimization solution and the learning-based unfolding strategy. Furthermore, as shown in~\cref{table:ablationStageNum}, we also analyze the most suitable stage number for our RUN framework. To achieve a balance between performance and efficiency, we set $K=4$.

\subsection{Further Analysis, Applications, and Meanings}

\noindent \textbf{Performance on small objects or multiple objects}. Small objects and multiple objects are challenging for lacking discriminative cues. To evaluate our performance on the two conditions, we filtered images from \textit{COD10K} that satisfy these criteria, resulting in $1,084$ images having concealed objects smaller than a quarter of the entire image and $186$ images with multiple concealed objects. As shown in~\cref{table:small-object,table:Multi-object}, while the performance of all methods declines, our approach consistently outperforms the competition.

\noindent \textbf{Performance on degraded COS scenarios}. To assess the impact of environmental degradation, we followed~\cite{he2023degradation} to simulate haze on concealed images in \textit{COD10K} and then evaluated the ability of the compared methods to resist degradation. As illustrated in~\cref{fig:ExperDegradation}, performance degrades as the haze concentration increases. However, our RUN demonstrates superior resilience to haze degradation, attributed to its multi-modality reversible modeling strategy.
% across both mask and RGB domains.
To enhance robustness, we replaced our reconstruction network $\mathcal{B}(\bigcdot)$ with a more complex network from CoRUN~\cite{fang2024real}, termed $\mathcal{B}_3(\bigcdot)$, which includes a pretrained dehazing model. This brought a novel unfolding network, RUN+, with $\mathcal{B}_3(\bigcdot)$ incorporating the pretrained model. \cref{fig:ExperDegradation} indicates integrating $\mathcal{B}_3(\bigcdot)$ enhances RUN’s robustness in resisting haze degradation. This underscores the potential of RUN in addressing degraded scenarios.
% involving environmental degradation.

\noindent \textbf{Potential applications of RUN}. 
% Here, we explore potential applications of our RUN framework. 
First, we test the effect of our RUN as a refiner,
% examine a straightforward integration of RUN with existing methods, 
specifically by initializing $\mathbf{M}_0$ with the results of existing methods. 
As shown in~\cref{table:RefinerRUN}, our approach can enhance the performance of SOTA methods without requiring retraining. Furthermore, we incorporate the core structures of existing methods into our RUN framework,
% by adapting them for use within the RSS module, 
followed by retraining the entire network. This integration yields even greater improvements, demonstrating that the unfolding framework can function as a plug-and-play solution to enhance the performance of existing methods. For example, as shown in~\cref{fig:Application}, we observe that while error predictions from FEDER influence FEDER-R, FEDER+ demonstrates better resilience to such errors. 
% Importantly, we dynamically adjust the batch size to ensure consistent training conditions across all configurations.

% 1bly,. to  serve as a plug-and-play framework; 2. serve as a  Notarefiner

\noindent \textbf{Meanings of our framework}.
Beyond introducing the deep unfolding network to high-level vision for the first time and enabling reversible modeling across both mask and RGB domains, the proposed RUN framework offers the potential to establish a \textit{unified vision strategy}.
% \textit{(1) Framework innovation}: This is the first attempt to introduce the deep unfolding network—a framework that balances interpretability and generalizability—to high-level vision tasks. RUN demonstrates its feasibility in challenging COS tasks, highlighting the potential of unfolding frameworks in advancing high-level vision applications.
% \textit{(2) Reversible modeling}: The framework enables reversible modeling of foreground and background in both the mask and RGB domains, effectively focusing the network on uncertain regions and reducing false-positive and false-negative outcomes. Experiments in~\cref{table:RefinerRUN,table:GeneraRUN} demonstrate that this framework can seamlessly integrate with existing SOTA COS methods, enhancing their performance.
% \textit{(3) Unified vision strategy}: 
By combining image segmentation and image reconstruction, our RUN introduces a novel approach to unifying low-level and high-level vision. 
As shown in~\cref{fig:discussion} in the appendix, unlike existing strategies, such as bi-level optimization~\cite{he2023HQG}, our unfolding-based combination strategy is underpinned by explicit theoretical guarantees with the two models better coupled. 
Moreover, as shown in~\cref{fig:ExperDegradation} RUN+, using more complex low-level vision algorithms results in a strong ability to resist complex degradation. 
% To further investigate this, we replaced our unfolding framework with bi-level optimization and discovered performance decreased by $1.84\%$ in $F_\beta$ and $2.28\%$ in $E_\phi$. 
This motivates further exploration of unfolding-based combination strategies to enhance high-level vision algorithms’ resistance to environmental degradation and imaging interference. Simultaneously, it promotes low-level vision algorithms by integrating deep semantic information and high-level guidance. Together, these advancements ensure practical applicability in both real-world high-level and low-level vision tasks.

\vspace{-1mm}
\section{Conclusions}\vspace{-1mm}
This paper proposes RUN to formulate the COS task as a foreground-background separation model. Its optimized solution is unfolded into a multistage network, where each stage comprises two reversible modules: SOFS and ROBE. SOFS applies the reversible strategy at the mask level and introduces RSS for non-local information extraction. ROBE employs a reconstruction network to address conflicting foreground and background regions in the RGB domain. Extensive experiments verify the superiority of RUN.

\newpage

%\section*{Impact Statement}
%This paper presents work whose goal is to advance the field of 
%Machine Learning. There are many potential societal consequences 
%of our work, none of which we feel must be specifically highlighted here.


\balance
\bibliography{example_paper}
\bibliographystyle{icml2025}

%%%%%%%%%%%%%%%%%%%%%%%%%%%%%%%%%%%%%%%%%%%%%%%%%%%%%%%%%%%%%%%%%%%%%%%%%%%%%%%
%%%%%%%%%%%%%%%%%%%%%%%%%%%%%%%%%%%%%%%%%%%%%%%%%%%%%%%%%%%%%%%%%%%%%%%%%%%%%%%
% APPENDIX
%%%%%%%%%%%%%%%%%%%%%%%%%%%%%%%%%%%%%%%%%%%%%%%%%%%%%%%%%%%%%%%%%%%%%%%%%%%%%%%
%%%%%%%%%%%%%%%%%%%%%%%%%%%%%%%%%%%%%%%%%%%%%%%%%%%%%%%%%%%%%%%%%%%%%%%%%%%%%%%
\newpage

\appendix
\onecolumn
\setcounter{figure}{0}
\renewcommand{\figurename}{Fig.}
\renewcommand{\thefigure}{S\arabic{figure}}
\setcounter{table}{0}
\renewcommand{\tablename}{Fig.}
\renewcommand{\thetable}{S\arabic{table}}
% \setcounter{algorithm}{1}
% % \renewcommand{\algorithmname}{Fig.}
\renewcommand{\thealgorithm}{S\arabic{algorithm}}
\setlength{\textfloatsep}{4pt}
% \begin{algorithm}[t]
%     %\setlength{\textfloatsep}{-5mm}
%     %\setlength{\floatsep}{0cm}
% 	\caption{Proposed RUN Framework.}
% 	\label{Alg:RUN}
% 	\textbf{Input}: concealed image $\mathbf{C}$, stage number $K$ \\
% 	\textbf{Output}: concealed object mask $\mathbf{M}_K$, concealed object edge $\mathbf{E}_K$, reconstructed concealed image $\hat{\mathbf{C}}_K$
% 	\begin{algorithmic}[1]
% 		\STATE Zero initialization for $\mathbf{M}_0$, $\mathbf{B}_0$
% 		% \STATE current epoch n $\gets 0$
% 		\FOR{each stage $k\in \left[1,K\right]$} 
%         % \textcolor{gray}{// Pseudo-label generation}
%             \STATE $\hat{\mathbf{M}}_k = \hat{\mathcal{M}}\left(\mathbf{B}_{k-1}, \mathbf{M}_{k-1}, \mathbf{M}_{k-2}, \mathbf{C}\right)$, 
%             \STATE $\mathbf{M}_k, \mathbf{E}_k  = \mathcal{M}( \mathbf{B}_{k-1}, \hat{\mathbf{M}}_k, \mathbf{C})$,
% 		\STATE $\hat{\mathbf{B}}_k =\hat{\mathcal{B}}\left(\mathbf{B}_{k-1}, \mathbf{M}_k, \mathbf{C}\right),$ 
%         \STATE ${\mathbf{B}}_k, \hat{\mathbf{C}}_k = {\mathcal{B}}\left(\hat{\mathbf{B}}_{k}, \mathbf{M}_k\right).$ 
% 		\ENDFOR
% 	\end{algorithmic}
% \end{algorithm}%\vspace{-3mm}
\begin{figure*}[h]
\setlength{\abovecaptionskip}{0cm}
	\centering
	\includegraphics[width=0.8\linewidth]{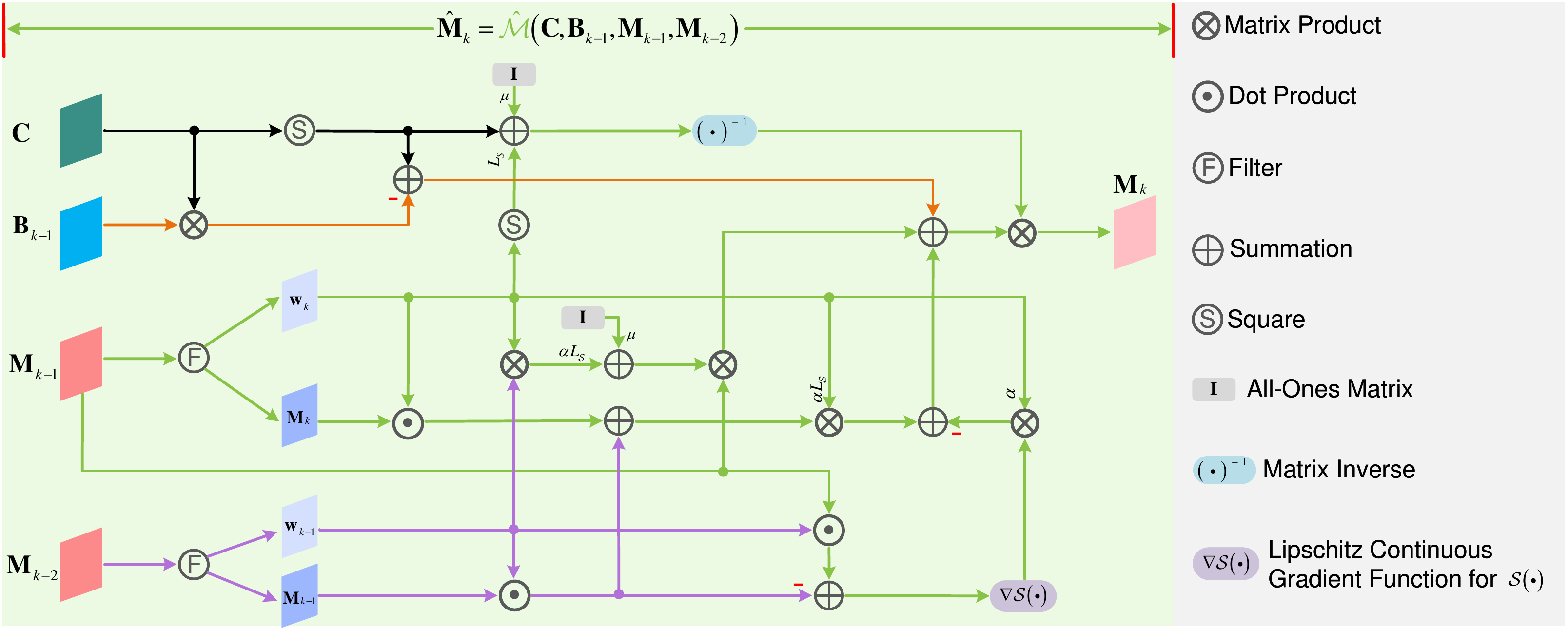}\vspace{-4mm}
	\caption{Details of $\hat{\mathcal{M}}(\cdot)$, where the connections in $\hat{\mathcal{M}}(\cdot)$ are derived strictly based on mathematical principles, thus enhancing interpretability.
    }
	\label{fig:SOFS_Details}
	\vspace{-5mm}
\end{figure*}

\section{Experiment}
\subsection{Datasets and metrics}\label{Sec:Dataset}
\noindent \textbf{Camouflaged object detection}. In this task, we follow the standard practice of SINet~\cite{fan2020camouflaged} and perform experiments on four datasets: \textit{CHAMELEON}~\cite{skurowski2018animal}, \textit{CAMO}~\cite{le2019anabranch}, \textit{COD10K}~\cite{fan2021concealed}, and \textit{NC4K}~\cite{lv2021simultaneously}. The \textit{CHAMELEON} dataset comprises 76 images, while the \textit{CAMO} dataset contains 1,250 images divided into 8 classes. The \textit{COD10K} dataset includes 5,066 images categorized into 10 super-classes, and \textit{NC4K} serves as the largest test set, with 4,121 images. For training, we use 1,000 images from \textit{CAMO} and 3,040 images from \textit{COD10K}. The remaining images from these two datasets, along with all images from the other datasets, constitute the test set.
To evaluate performance, we employ four widely-used metrics: mean absolute error ($M$), adaptive F-measure ($F_\beta$)~\cite{margolin2014evaluate}, mean E-measure ($E_\phi$)~\cite{fan2021cognitive}, and structure measure ($S_\alpha$)~\cite{fan2017structure}. Superior performance is indicated by lower values of $M$ and higher values of $F_\beta$, $E_\phi$, and $S_\alpha$.

\noindent \textbf{Medical concealed object segmentation}. We evaluate the performance of our method on two specific tasks: polyp image segmentation and medical tubular object segmentation.
For polyp image segmentation, we utilize two benchmarks: \textit{CVC-ColonDB}~\cite{tajbakhsh2015automated} and \textit{ETIS}~\cite{silva2014toward}. The training protocol follows the setup of LSSNet~\cite{wang2024lssnet}. Quantitative evaluation is conducted using three commonly adopted metrics: mean Dice (mDice), mean Intersection over Union (mIoU), and structure measure ($S_\alpha$), where higher values indicate better performance.
For medical tubular object segmentation, we evaluate our method on the \textit{DRIVE}\footnote{http://www.isi.uu.nl/Research/Databases/DRIVE/} and \textit{CORN}~\cite{ma2021structure} datasets, with training and inference conducted separately for each dataset. For the \textit{DRIVE} dataset, training and inference adhere to the dataset’s predefined splits. For the \textit{CORN} dataset, the last $70\%$ of the data is used for training, while the first $30\%$ serves as the test set. Following DSCNet~\cite{qi2023dynamic}, we employ three evaluation metrics: mDice, area under the ROC curve (AUC), and sensitivity (SEN), with higher values reflecting better performance.
To ensure a fair comparison with state-of-the-art medical concealed object segmentation methods, which predominantly utilize transformer-based encoders, we adopt PVT V2 as the backbone for our encoder.

% Two specific tasks are selected, including polyp image segmentation and medical tubular object segmentation. For the polyp image segmentation task, we evaluate the performance of our method on two benchmarks: \textit{CVC-ColonDB}~\cite{tajbakhsh2015automated} and \textit{ETIS}~\cite{silva2014toward}. Our training rule follows the setup of LSSNet~\cite{wang2024lssnet}. For quantitative evaluation, we adopt three commonly used metrics: mean Dice (mDice), mean Intersection over Union (mIoU), and structure measure ($S_\alpha$). Notably, higher values are preferred for these metrics. For the medical tubular object segmentation task, we evaluate our method on the \textit{DRIVE} and \textit{CORN} datasets. The two datasets are trained and inference separately. In the \textit{DRIVE} dataset, we train and inference according to the divided datasets. For the \textit{CORN} dataset, We use the last $70\%$ of the data to train the network and the first $30\%$ of the data to test the training effect. Following DSCNet~\cite{qi2023dynamic}, three metrics are used, including mDice, the area under the ROC curve (AUC), and sensitivity (SEN). Among these three metrics, higher values indicate better performance. Given that state-of-the-art Medical concealed object segmentation methods typically utilize transformer-based encoders, we employ PVT V2 as the backbone for our encoder to ensure a fair comparison.

\noindent \textbf{Transparent object detection}. For a fair comparison, we use PVT V2 as our default backbone and conduct experiments on two datasets: \textit{GDD} \cite{mei2020don} and \textit{GSD} \cite{lin2021rich}. The training set consists of 2,980 images from \textit{GDD} and 3,202 images from \textit{GSD}, while the remaining images are reserved for inference. Consistent with GDNet-B~\cite{mei2022large}, we evaluate performance using several metrics, including mIoU, and maximum F-measure ($F_\beta^{max}$).
% , and balanced error rate (BER). 
Superior performance is indicated by lower values for $M$, or higher values for mIoU and $F_\beta^{max}$.

\noindent \textbf{Concealed defect detection}. In this task, we utilize PVT V2 as the default backbone. Consistent with established practices, we evaluate the generalization capacity of our RUN framework on the concealed defect detection task. Specifically, we use the model trained on the COD task to segment concealed objects in the \textit{CDS2K} dataset~\cite{fan2023advances}. Eight evaluation metrics are employed, where higher values indicate better performance for all metrics except MAE, for which lower values are preferred.

\noindent \textbf{Salient object detection}. We evaluate the performance of our method on five widely used benchmark datasets: \textit{DUT-OMRON} \cite{yang2013saliency}, \textit{DUTS-test} \cite{wang2017learning}, \textit{ECSSD} \cite{yan2013hierarchical}, \textit{HKU-IS} \cite{li2015visual}, and \textit{PASCAL-S} \cite{li2014secrets}. The \textit{DUTS} dataset contains 10,553 training images and 5,019 test images, referred to as \textit{DUTS-test}. The \textit{DUT-OMRON} and \textit{ECSSD} datasets include 5,168 and 1,000 images, respectively. The \textit{HKU-IS} dataset consists of 4,447 images featuring multiple foreground salient objects, while \textit{PASCAL-S} comprises 850 images derived from the \textit{PASCAL VOC 2010} dataset \cite{everingham2010pascal}. For evaluation, we use the same metrics applied in COD.

\subsection{Generalization on concealed defect detection}\label{Sec:CDD}
We evaluate the generalization capability of our RUN framework in the concealed defect detection task by directly using the model trained on the COD task. Detailed information about the experimental setup can be found in~\cref{Sec:Dataset}. As shown in~\cref{table:CDDQuanti}, our method achieves superior performance compared to existing state-of-the-art approaches, further highlighting the advancements and effectiveness of the RUN framework.

\subsection{Generalization on salient object detection}\label{Sec:SOD}
% \noindent \textbf{Applications on salient object detection}. 
We evaluate the generalization of our method in salient object detection. Details regarding the training configurations and datasets are provided in \cref{Sec:Dataset}. As shown in~\cref{table:SODQuanti}, our method outperforms existing state-of-the-art approaches, achieving a leading position. These results highlight the superiority of our approach and underscore the potential of unfolding-based frameworks for high-level vision tasks.

\section{Limitations and Future Work}
As illustrated in~\cref{fig:ExperDegradation}, our RUN model, like other advanced methods, exhibits instability in degraded scenarios. This is primarily because environmental degradation exacerbates the challenge of extracting subtle discriminative information, bringing difficulties in capturing concealed objects. However, RUN+ demonstrates robustness to haze degradation, underscoring the potential of the RUN-based framework for addressing scenarios involving environmental degradation.

\begin{table*}[t]
 \begin{minipage}[c]{\textwidth}
	\centering
	\setlength{\abovecaptionskip}{0cm}
	\caption{Restuls on concealed defect detection.  
    % We employ PVT V2 as our backbone.
    } \label{table:CDDQuanti}
	\resizebox{1\columnwidth}{!}{
		\setlength{\tabcolsep}{7mm}
		\begin{tabular}{l|cccccccc}
        \toprule 
Methods    & {\cellcolor{gray!40}$S_\alpha$~$\uparrow$} & {\cellcolor{gray!40}$M$~$\downarrow$} & {\cellcolor{gray!40}$E_\phi^{ad}$~$\uparrow$}  & {\cellcolor{gray!40}$E_\phi$~$\uparrow$} & {\cellcolor{gray!40}$E_\phi^{max}$~$\uparrow$} & {\cellcolor{gray!40}$F_\beta$~$\uparrow$} & {\cellcolor{gray!40}$F_\beta^{mean}$~$\uparrow$} & {\cellcolor{gray!40}$F_\beta^{max}$~$\uparrow$} \\ \midrule
SINet V2~\cite{fan2021concealed}   & 0.551 & 0.102 & 0.509  & 0.567  & {\color[HTML]{00B0F0} \textbf{0.597}}  & 0.223  & 0.248  & 0.258  \\
HitNet~\cite{hu2022high}     & 0.563 & 0.118 & 0.574  & 0.564  & 0.570  & {\color[HTML]{00B0F0} \textbf{0.298}}  & 0.298  & 0.299  \\
% DGNet      & 0.578 & 0.258 & 0.089 & 0.552  & 0.569  & 0.579  & 0.274  & 0.291  & 0.297  \\
CamoFormer~\cite{yin2024camoformer} & {\color[HTML]{00B0F0} \textbf{0.589}} & {\color[HTML]{00B0F0} \textbf{0.100}} & {\color[HTML]{00B0F0} \textbf{0.590}}  & {\color[HTML]{00B0F0} \textbf{0.588}}  & 0.596  & {\color[HTML]{FF0000} \textbf{0.330}}  & {\color[HTML]{FF0000} \textbf{0.329}}  & {\color[HTML]{FF0000} \textbf{0.339}}  \\
OAFormer~\cite{yang2023oaformer}   & 0.541 & 0.121 & 0.479  & 0.535  & 0.591  & 0.216  & 0.239  & 0.252  \\
\rowcolor{c2!20} RUN (Ours) & {\color[HTML]{FF0000} \textbf{0.590}} &  {\color[HTML]{FF0000} \textbf{0.068}} & {\color[HTML]{FF0000} \textbf{0.601}} & {\color[HTML]{FF0000} \textbf{0.595}} & {\color[HTML]{FF0000} \textbf{0.611}}  & {\color[HTML]{00B0F0} \textbf{0.298}}  & {\color[HTML]{00B0F0} \textbf{0.299}}  & {\color[HTML]{00B0F0} \textbf{0.303}}  \\
\bottomrule
\end{tabular}}
	% \vspace{-0.6cm}
  \end{minipage}
\\
 \begin{minipage}[c]{\textwidth}
	\centering
	\setlength{\abovecaptionskip}{0cm}
	\caption{Restuls on salient object detection.  
    % We employ PVT V2 as our backbone.
    } \label{table:SODQuanti}
	\resizebox{\columnwidth}{!}{
		\setlength{\tabcolsep}{0.8mm}
		\begin{tabular}{l|cccc|cccc|cccc|cccc|cccc} 
			\toprule 
			\multicolumn{1}{l|}{}& \multicolumn{4}{c|}{\textit{DUT-OMRON} }& \multicolumn{4}{c|}{\textit{DUTS-test} }& \multicolumn{4}{c|}{\textit{ECSSD} }& \multicolumn{4}{c|}{\textit{HKU-IS} }& \multicolumn{4}{c}{\textit{PASCAL-S} }\\ \cline{2-21}
			\multicolumn{1}{l|}{\multirow{-2}{*}{Methods}} & {\cellcolor{gray!40}$M$~$\downarrow$} &{\cellcolor{gray!40}$F_\beta$~$\uparrow$} &{\cellcolor{gray!40}$E_\phi$~$\uparrow$} & \multicolumn{1}{c|}{\cellcolor{gray!40}$S_\alpha$~$\uparrow$}& {\cellcolor{gray!40}$M$~$\downarrow$} &{\cellcolor{gray!40}$F_\beta$~$\uparrow$} &{\cellcolor{gray!40}$E_\phi$~$\uparrow$} & \multicolumn{1}{c|}{\cellcolor{gray!40}$S_\alpha$~$\uparrow$}& {\cellcolor{gray!40}$M$~$\downarrow$} &{\cellcolor{gray!40}$F_\beta$~$\uparrow$} &{\cellcolor{gray!40}$E_\phi$~$\uparrow$} & \multicolumn{1}{c|}{\cellcolor{gray!40}$S_\alpha$~$\uparrow$}& {\cellcolor{gray!40}$M$~$\downarrow$} &{\cellcolor{gray!40}$F_\beta$~$\uparrow$} &{\cellcolor{gray!40}$E_\phi$~$\uparrow$} & \multicolumn{1}{c|}{\cellcolor{gray!40}$S_\alpha$~$\uparrow$}& {\cellcolor{gray!40}$M$~$\downarrow$} &{\cellcolor{gray!40}$F_\beta$~$\uparrow$} &{\cellcolor{gray!40}$E_\phi$~$\uparrow$} & \multicolumn{1}{c}{\cellcolor{gray!40}$S_\alpha$~$\uparrow$}\\ \midrule
% CPD~\cite{wu2019cascaded} & 0.056 & 0.753 & 0.873 & 0.825  & 0.043 & 0.813 & 0.901 & 0.869 & 0.037 & 0.892 & 0.949 & 0.918 & 0.034 & 0.862 & 0.950 & 0.906 & 0.071 & 0.736 & 0.887 & 0.848   \\
% MINet~\cite{pang2020multi} & 0.056 & 0.757 & 0.873 & 0.833 & 0.037 & 0.835 & 0.913 & 0.884 & 0.034 & 0.866 & 0.953 & 0.925 & 0.029 & 0.847 & 0.960 & 0.918 & 0.064 & 0.807 & 0.898 & 0.856	\\
VST~\cite{liu2021visual} & 0.058 & 0.755 & 0.871 &0.850 & 0.037 &0.828&0.919 & 0.896 & 0.033 & 0.910 & 0.951 & 0.932 & 0.029 & 0.897 & 0.952 & 0.928 & 0.061 & 0.816 & 0.902 & 0.872  \\
ICON-P~\cite{zhuge2022salient} & 0.047 & {\color[HTML]{FF0000} \textbf{0.793}} & {\color[HTML]{FF0000} \textbf{0.896}} & {\color[HTML]{00B0F0} \textbf{0.865}} & {\color[HTML]{FF0000} \textbf{0.022}} & {\color[HTML]{00B0F0} \textbf{0.882}} & {\color[HTML]{00B0F0} \textbf{0.950}} & {\color[HTML]{FF0000} \textbf{0.917}} & {\color[HTML]{00B0F0} \textbf{0.024}} & {\color[HTML]{00B0F0} \textbf{0.933}} & 0.964&0.940 & {\color[HTML]{FF0000} \textbf{0.022}} & {\color[HTML]{00B0F0} \textbf{0.925}} & 0.967 & {\color[HTML]{00B0F0} \textbf{0.935}} & {\color[HTML]{FF0000} \textbf{0.051}} & {\color[HTML]{FF0000} \textbf{0.847}} & {\color[HTML]{00B0F0} \textbf{0.921}} &  {\color[HTML]{00B0F0} \textbf{0.882}} \\
PGNet~\cite{xie2022pyramid} & {\color[HTML]{FF0000} \textbf{0.045}} & 0.767 & 0.887 & 0.855 & 0.027 & 0.851 & 0.922 & 0.911 & {\color[HTML]{00B0F0} \textbf{0.024}} & 0.920 & 0.955 & 0.932 & 0.024 & 0.912 & 0.958 & 0.934 & 0.052 & 0.838 & 0.912 & 0.875 \\
MENet~\cite{wang2023pixels} & {\color[HTML]{FF0000} \textbf{0.045}} & 0.782 & 0.891 & 0.849 & 0.028 & 
0.860 & 0.937 & 0.905& 0.033 & 0.906 & 0.954 & 0.928 &0.023 & 0.910 & 0.966 & 0.927 & 0.054 & 0.838 & 0.913 & 0.872\\
RMFormer~\cite{deng2023recurrent} & 0.049 & 0.775 & 0.892 & 0.862 & 0.030 & 0.850 & 0.928 & 0.907 & 0.028 & 0.917 & 0.957 & 0.933 & 0.024 & 0.908 & 0.960 & 0.930 &0.057 & 0.827 & 0.909 & 0.869 \\
GPONet~\cite{yi2024gponet} & {\color[HTML]{FF0000} \textbf{0.045}} & 0.788 & 0.889 & {\color[HTML]{00B0F0} \textbf{0.865}} & 0.027 & 0.858 & 0.937 & 0.912 & 0.025 & 0.925 & 0.964 & {\color[HTML]{FF0000} \textbf{0.942}} & 0.023 & 0.918 & 0.962 & {\color[HTML]{FF0000} \textbf{0.936}} & 0.055 & 0.836 & 0.908 & 0.870   \\
% VSCode~\cite{luo2024vscode} &---&---& 0.912 & 0.877 &---&---& 0.960 & 0.926 &---&---& 0.974 & 0.949 &---&---& 0.974 & 0.940 &---&---& 0.904 & 0.887  \\
VST-T++~\cite{liu2024vst++} &0.046&0.778& 0.892 & 0.853 &0.028 & 0.869& 0.943 & 0.901 &0.025&0.930& {\color[HTML]{00B0F0} \textbf{0.968}} & 0.937 & 0.024 & 0.919 & {\color[HTML]{00B0F0} \textbf{0.968}} & 0.930 & {\color[HTML]{FF0000} \textbf{0.051}} & 0.841 & 0.901 & 0.878  \\
\rowcolor{c2!20}RUN (Ours) & {\color[HTML]{FF0000} \textbf{0.045}}  & 
 {\color[HTML]{FF0000} \textbf{0.793}} &{\color[HTML]{00B0F0} \textbf{0.893}}  & {\color[HTML]{FF0000} \textbf{0.867}}  & {\color[HTML]{FF0000} \textbf{0.022}} & {\color[HTML]{FF0000} \textbf{0.886}} & {\color[HTML]{FF0000} \textbf{0.953}} & {\color[HTML]{00B0F0} \textbf{0.916}} & {\color[HTML]{FF0000} \textbf{0.023}} & {\color[HTML]{FF0000} \textbf{0.935}} & {\color[HTML]{FF0000} \textbf{0.971}} & {\color[HTML]{00B0F0} \textbf{0.941}} & {\color[HTML]{FF0000} \textbf{0.022}} & {\color[HTML]{FF0000} \textbf{0.927}} & {\color[HTML]{FF0000} \textbf{0.970}} & 0.934 & {\color[HTML]{FF0000} \textbf{0.051}} & {\color[HTML]{00B0F0} \textbf{0.843}} & {\color[HTML]{FF0000} \textbf{0.925}}  & {\color[HTML]{FF0000} \textbf{0.883}} \\
   \bottomrule 
	\end{tabular}}
	% \vspace{-0.2cm}
  \end{minipage}
\end{table*}

\begin{figure}[t]
\setlength{\abovecaptionskip}{0cm}
	\centering
	\includegraphics[width=0.5\linewidth]{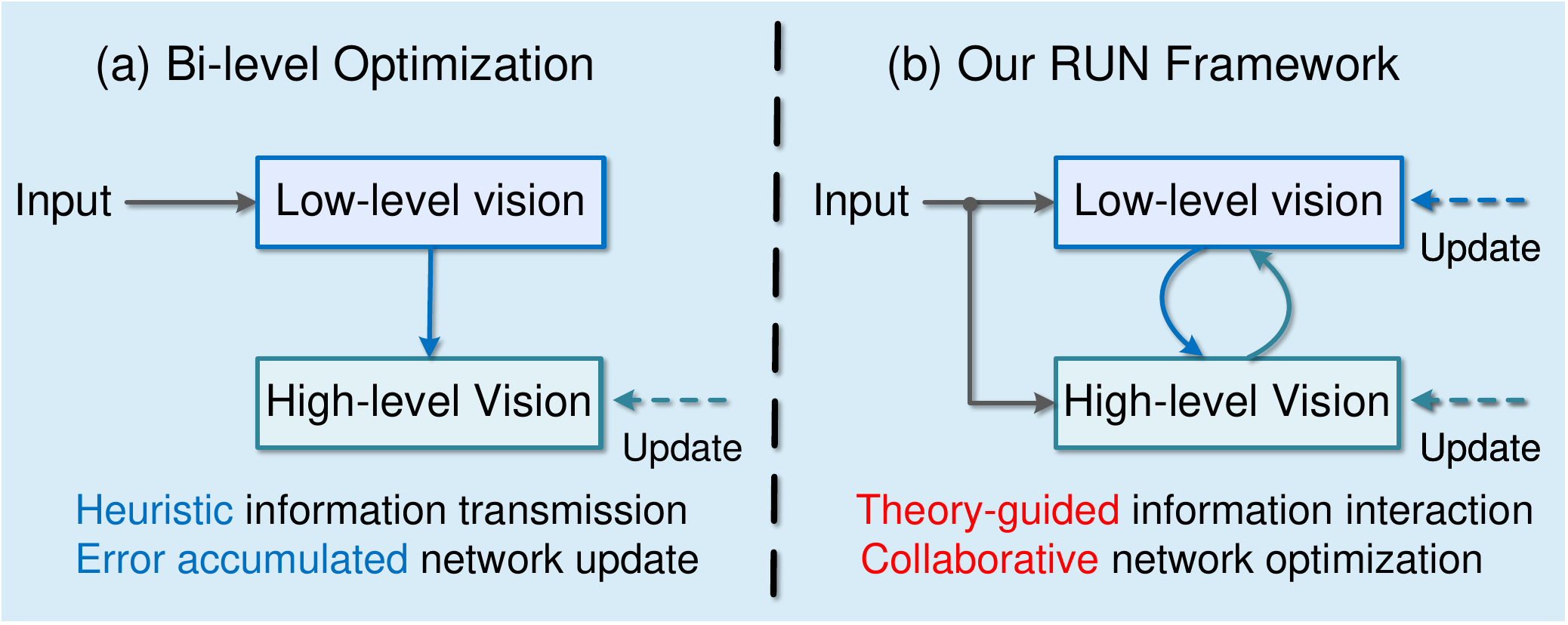 }\vspace{-4mm}
	\caption{Comparison between bi-level optimization and our RUN.}
	\label{fig:discussion}
	% \vspace{-1mm}
\end{figure}
Future work will focus on integrating the RUN model with more advanced low-level vision algorithms to address increasingly complex scenarios involving diverse types of degradation, such as low light, blur, and noise. Additionally, incorporating large-scale algorithms into the unfolding-based multi-stage framework introduces significant computational and storage demands. Developing strategies to effectively integrate degradation-resistant models within this framework remains an important research direction.

Furthermore, as this work represents a pioneering effort in applying deep unfolding networks (DUNs) to high-level vision tasks, it opens the door for the development of more DUN-based algorithms in this domain. These future approaches are expected to better balance interpretability and generalizability, further advancing high-level vision tasks. Additionally, establishing an unfolding-based high-resolution COS method~\cite{zheng2024bilateral} is also our goal.

% As shown in~\cref{fig:ExperDegradation}, our RUN model, along with other cutting-edge methods, performs unstably in degraded scenarios. This is because degraded environments further influence the extraction of the subtle discriminative information and thus lead to the failure capture of concealed objects. Nevertheless, RUN+ exhibits robustness to haze degradation, indicating that the RUN-based framework has the potential to address scenarios involving environmental degradation. Our future work lies in combining our RUN model with more complex low-level vision algorithms and solving more sophisticated scenarios with diverse kinds of degradation, such as low light, blur, and noise.
% Additionally, simply integrating a large-scale algorithm into the unfolding-based multi-stage framework can bring huge computational and storage requirements. In this case,  how to effectively combine the degradation-resistant model is a worth-studying direction. 
% Furthermore, given that this is a pioneering work of applying deep unfolding networks into the field of high-level vision tasks, more DUN-based algorithms are expected to be proposed in high-level vision tasks, balancing interpretability and generalizability.

%%%%%%%%%%%%%%%%%%%%%%%%%%%%%%%%%%%%%%%%%%%%%%%%%%%%%%%%%%%%%%%%%%%%%%%%%%%%%%%
%%%%%%%%%%%%%%%%%%%%%%%%%%%%%%%%%%%%%%%%%%%%%%%%%%%%%%%%%%%%%%%%%%%%%%%%%%%%%%%

\end{document}